\documentclass{article} 
\usepackage{iclr2027_conference,times}

\usepackage{amsmath,amsfonts,bm}

\def\eqref#1{equation~\ref{#1}}

\def\1{\bm{1}}

\DeclareMathAlphabet{\mathsfit}{\encodingdefault}{\sfdefault}{m}{sl}
\SetMathAlphabet{\mathsfit}{bold}{\encodingdefault}{\sfdefault}{bx}{n}

\usepackage[hypertexnames=false]{hyperref}
\usepackage{url}
\usepackage{titletoc}
\usepackage{float}
\usepackage{needspace}
\title{LabCompiler: A Framework for Verifiable Human-Machine Laboratory Execution from Imperfect Protocols}

\author{%
\makebox[\textwidth][l]{%
\begin{tabular}{@{}l@{}}
\textbf{Jingyuan Li}\textsuperscript{1,2*},
\textbf{Dongzhan Zhou}\textsuperscript{2*},
\textbf{Minting Pan}\textsuperscript{2\ensuremath{\dagger}},
\textbf{Yujie Liu}\textsuperscript{2,3},
\textbf{Yuan Gao}\textsuperscript{4},
\textbf{Yankai Jiang}\textsuperscript{2}, \\
\textbf{Xinyu Gu}\textsuperscript{2},
\textbf{Zhe Liu}\textsuperscript{5},
\textbf{Yang Li}\textsuperscript{1},
\textbf{Bo Zhang}\textsuperscript{2},
\textbf{Lei Bai}\textsuperscript{2},
\textbf{Tong Si}\textsuperscript{4\ensuremath{\dagger}},
\textbf{Weinan Zhang}\textsuperscript{1\ensuremath{\dagger}}
\\[4pt]
\textmd{%
\textsuperscript{1}Shanghai Jiao Tong University
\qquad
\textsuperscript{2}Shanghai Artificial Intelligence Laboratory
}
\\[-1pt]
\textmd{%
\textsuperscript{3}Fudan University
\qquad
\textsuperscript{4}Shenzhen Institutes of Advanced Technology
}
\\[-1pt]
\textmd{%
\textsuperscript{5}East China University of Science and Technology
}
\\[2pt]
\textmd{%
\textsuperscript{*}Equal contribution.
\qquad
\textsuperscript{\ensuremath{\dagger}}Corresponding authors.
}
\end{tabular}%
}%
}

\usepackage{enumitem}
\usepackage{amssymb}
\usepackage{booktabs}
\usepackage{tabularx}
\usepackage{array}
\usepackage[table]{xcolor}
\usepackage{makecell}
\usepackage{multirow}
\usepackage{graphicx} 
\usepackage{longtable}

\definecolor{MyDarkBlue}{rgb}{0,0.5,1}
\definecolor{MyDarkGreen}{rgb}{0.02,0.6,0.02}
\definecolor{MyDarkRed}{rgb}{0.8,0.02,0.02}
\definecolor{MyDarkOrange}{rgb}{0.40,0.2,0.02}
\definecolor{MyPurple}{RGB}{111,0,255}
\definecolor{MyRed}{rgb}{1.0,0.0,0.0}
\definecolor{MyGold}{rgb}{0.75,0.6,0.12}
\definecolor{MyDarkgray}{rgb}{0.66, 0.66, 0.66}

\usepackage{listings}

\lstdefinestyle{json}{
    basicstyle=\ttfamily\footnotesize,
    frame=single,
    breaklines=true,
    breakatwhitespace=false,
    columns=fullflexible,
    keepspaces=true,
    showstringspaces=false,
    tabsize=2,
    xleftmargin=2pt,
    xrightmargin=2pt,
    framexleftmargin=2pt,
    framexrightmargin=2pt,
    rulecolor=\color{black!30},
    backgroundcolor=\color{black!2},
    captionpos=b
}

\newcommand{\model}{LabCompiler}

\iclrfinalcopy 
\begin{document}

\maketitle
\lhead{}

\begin{abstract}

Closing the loop of AI-driven science requires reliably translating natural-language protocols into executable laboratory workflows despite protocol defects and heterogeneous hardware. Existing approaches often couple scientific interpretation with device-specific code generation, making errors difficult to localize and limiting reuse across laboratory platforms. We introduce \textbf{\model{}}, a robust protocol-to-execution framework that transforms imperfect natural-language protocols into verifiable human-machine workflows. 
Its protocol firewall diagnoses and repairs defects before compilation while preserving unaffected content, guided by troubleshooting knowledge accumulated through perturbation-guided self-evolution. A two-level, device-agnostic graph representation separates experimental semantics from hardware-specific implementation and supports capability-based device assignment with coordinated human execution. We further introduce a protocol recovery benchmark covering four controlled perturbation types and naturally occurring errors collected from protocols written by iGEM participants.
Experiments show that \model{} outperforms frontier LLMs in protocol error correction while enabling coordinated human-machine execution across heterogeneous, partially automated laboratory settings.

\end{abstract}
\section{Introduction}
\label{sec:intro}

Artificial intelligence for science (AI4S) is moving toward closed-loop experimentation. Recent systems can synthesize literature~\citep{qiu2025completing}, formulate hypotheses~\citep{baek2025researchagent}, and design experimental procedures~\citep{huang2025automated}, while self-driving laboratories and robotic platforms are beginning to automate experimental execution and iterative optimization~\citep{boiko2023autonomous}. Connecting these capabilities requires translating natural-language protocols into actions that can be carried out in a particular laboratory. Researchers usually perform this translation, resolving implicit details and deciding how the available instruments should be used. When this work depends on individual interpretation, the same protocol may be executed differently by different operators. A standardized interface between experimental descriptions and laboratory actions would allow this translation to be reproduced and audited, making laboratory automation easier to scale.

Coding agents provide a natural foundation for such an interface, as they can generate programs, invoke external tools, and iteratively revise code based on execution feedback~\citep{wang2024executable,yang2024swe}. Recent work has explored the use of coding agents to translate experimental protocols into machine-executable scripts~\citep{inagaki2023llms,gao2025autonomous,jiang2026self}.
Nevertheless, they still face three fundamental challenges.
First, the protocol itself may be imperfect, containing incorrect parameters, missing steps, or ambiguous instructions. Direct compilation may therefore produce valid code that faithfully implements a scientifically invalid experiment. 
Second, direct protocol-to-code generation entangles scientific interpretation with device-specific implementation, making failures difficult to localize and limiting reuse across heterogeneous hardware.
Third, many protocols are only partially automatable with available devices and require human manipulation or judgment to complete the experiment. 

\begin{figure*}[t]
\begin{center}
\centerline{\includegraphics[width=\linewidth]{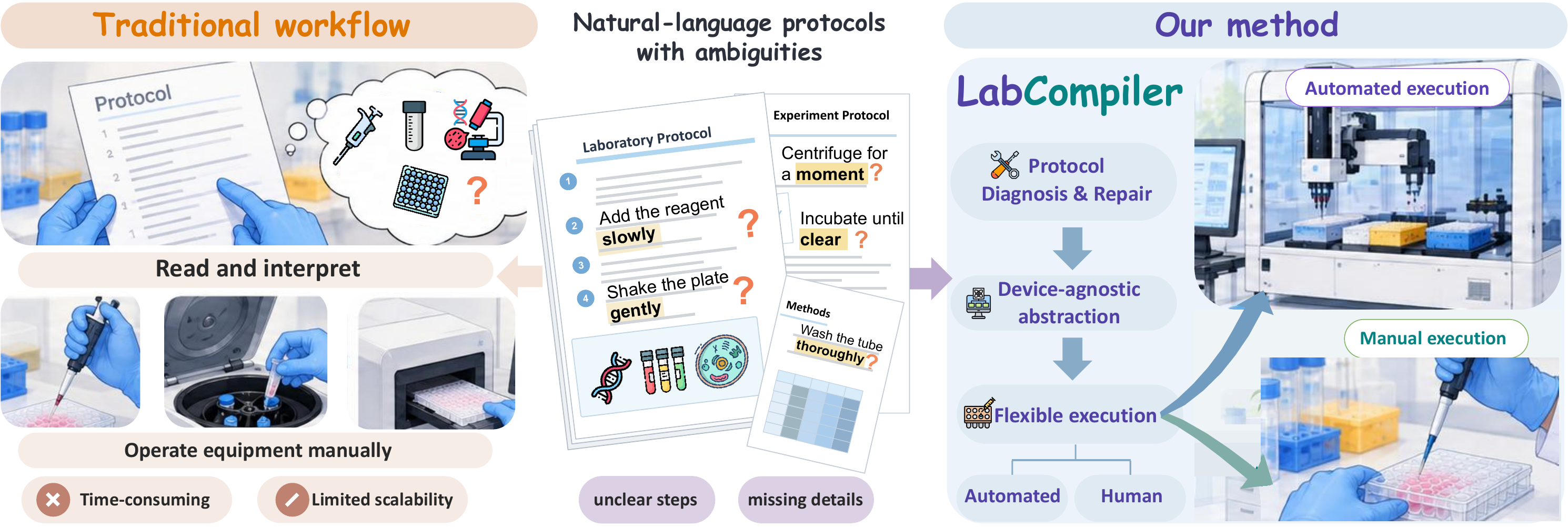}}
\vspace{-5pt}
\caption{Overview of LabCompiler. Unlike traditional workflows that rely on manual interpretation and operation, LabCompiler transforms ambiguous natural-language protocols into human-machine workflows through diagnosis and repair, device-agnostic abstraction, and hybrid execution.}
\label{fig:intro}
\end{center}
\vspace{-25pt}
\end{figure*}

We introduce \model{}, a robust protocol-to-execution framework that addresses these challenges through a staged ``diagnose-represent-execute'' architecture, as illustrated in Figure~\ref{fig:intro}. Its \emph{protocol firewall} diagnoses and repairs defects before compilation while constraining unnecessary changes to valid experimental content. The firewall incorporates \emph{perturbations-guided self-evolution}, through which diagnosis and repair outcomes on perturbed training protocols are examined against verified references and distilled into reusable troubleshooting knowledge. Accumulated in an external memory, this knowledge guides subsequent screening.

The repaired protocol is compiled into a \emph{Semantic Experiment Graph} that captures operations and their control and material dependencies, then deterministically normalized into a \emph{Canonical Action Graph} of standardized actions. This device-agnostic representation decouples experimental semantics from hardware-specific implementation. Related actions are grouped into execution units and assigned to compatible devices when possible, while unsupported operations remain human-execution steps. Automated units are grounded in physical resources, translated into device-specific code, and checked against device and resource constraints, with simulation used when available. The shared representation also allows execution assignments to adapt as laboratory capabilities expand.

We construct a benchmark for protocol recovery, which requires systems to identify and repair defects without predefined corrective options while preserving unaffected experimental information. 
Unlike LAB-Bench~\citep{laurent2024lab}, which uses multiple-choice questions for protocol troubleshooting, our benchmark evaluates protocol recovery by requiring systems to directly identify and repair defects in imperfect protocols. 
It combines structured perturbations of validated public protocols with naturally occurring errors curated from protocols written by iGEM competition teams and annotated by domain experts, enabling us to evaluate specific failure modes and examine whether recovery extends to errors encountered in practice.
Our contributions are as following:
\begin{itemize}[leftmargin=*]
    \item We introduce \model{}, a staged protocol-to-execution framework that diagnoses protocol defects before compilation, reducing the propagation of protocol errors into downstream execution.

    \item We develop a two-level device-agnostic experiment representation and capability-grounded execution mechanism that separates scientific semantics from hardware-specific implementation while supporting coordinated machine-human workflows.
    
    \item We introduce a protocol correction benchmark and correction-guided self-evolution mechanism for evaluating and improving protocol diagnosis and repair.
\end{itemize}

\section{Related Work}
\label{sec:related}

\paragraph{Agentic Laboratory Automation.}
Autonomous laboratories are emerging as infrastructure for closing the loop from scientific reasoning to physical experimentation. Early platforms commonly encoded task-specific procedures in fixed orchestration workflows, making adaptation to new protocols or instruments costly~\citep{macleod2020self,seifrid2022autonomous,lo2024review,rapp2024self}.
Recent advances in large language models and coding agents have made this interface more flexible, enabling goal-level experimental instructions to be translated into executable robotic scripts~\citep{ruan2024automatic,gao2025autonomous,jiang2026self}.
However, many such approaches couple scientific interpretation with device-specific execution within the same generation process. Thus, experimental semantics are tied to particular hardware realizations, so transferring protocols across instruments often requires regenerating or revising backend-specific implementations.
Complementary work has introduced machine-readable and hardware-agnostic representations, including XDL~\citep{mehr2020universal}, BPL~\citep{song2026towards}, and LabOP~\citep{bartley2023building}, to standardize experimental procedures and improve interoperability across laboratory platforms.
Building on this direction, \model{} introduces a two-level, device-agnostic experimental representation that explicitly separates protocol understanding from hardware realization, allowing the same experimental intent to be consistently mapped to heterogeneous laboratory platforms.

\vspace{-5pt}
\paragraph{Protocol Diagnosis and Repair.}

Protocols used in practice may contain errors, ambiguities, or missing information, making direct execution unsafe.
This has motivated recent agentic laboratory systems to incorporate protocol validation and refinement before robotic grounding or code generation, treating scientific correctness as a prerequisite for physical execution~\citep{liu2026bioproagent, liu2026protoact,jiang2026self}.
In parallel, recent benchmarks such as LAB-Bench and BioProBench have begun assessing whether language models can identify and correct errors in biological protocols~\citep{laurent2024lab,laurent2026labbench2,liu2026bioprobench,ivanov2024biolp}.
Our benchmark evaluates protocol recovery under two complementary settings: controlled errors introduced into validated protocols, and naturally occurring errors identified in protocols written by iGEM participants and corrected by domain experts. This design supports both controlled analysis of specific failure modes and evaluation on errors encountered in practice.
Additionally, \model{} couples an iterative diagnose-repair-verify loop with a perturbation-guided self-evolving diagnosis mechanism, distilling resolved cases into reusable troubleshooting knowledge for subsequent protocol diagnosis and repair.

\section{\model{}}
\label{sec:method}

\begin{figure*}[t]
\begin{center}
\centerline{\includegraphics[width=\linewidth]{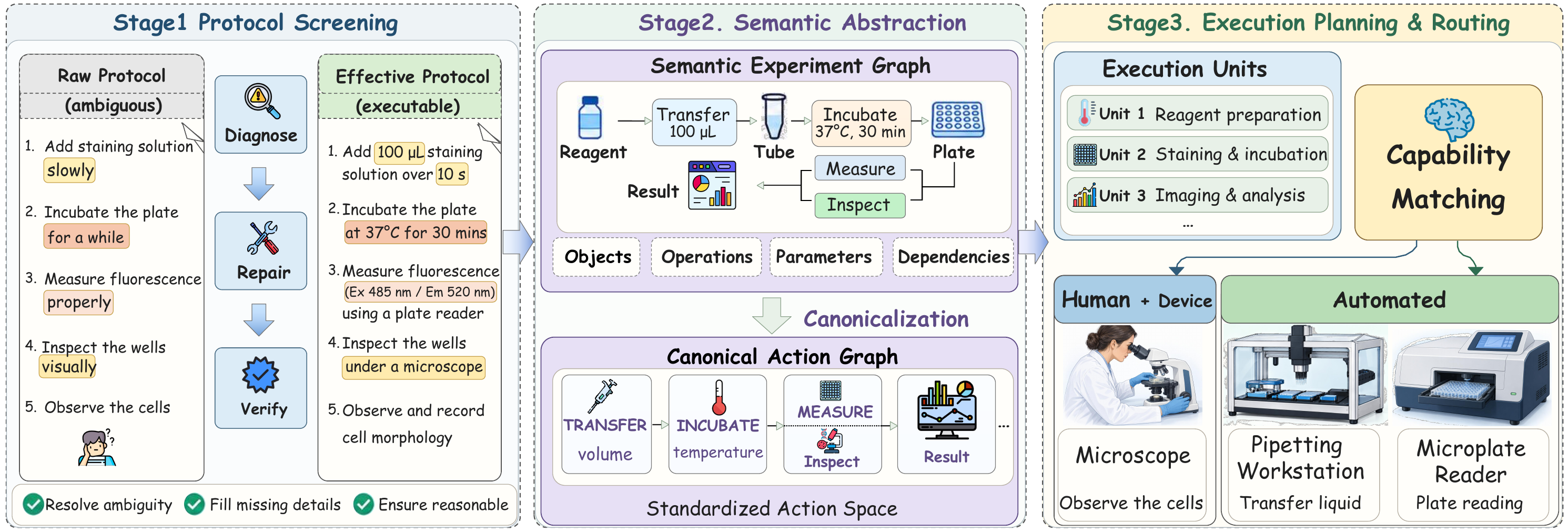}}
\vspace{-5pt}
\caption{Overview of LabCompiler. It first screens and repairs ambiguous natural-language protocols, then converts them into device-agnostic semantic and canonical representations. The resulting actions are grouped into execution units and capability-matched to human or automated execution.}
\label{fig:framework}
\end{center}
\vspace{-25pt}
\end{figure*}

\model{} transforms a natural-language experimental protocol into a verifiable laboratory workflow through protocol verification, device-agnostic semantic compilation, and capability-aware human-machine execution.
Formally, the input is $\mathcal I=(g,\mathcal P)$, where $g$ denotes the experimental goal and $\mathcal P$ the ordered procedures. \model{} first detects and repairs protocol-level defects, yielding a verified protocol $\widetilde{\mathcal P}$, then maps it into a device-agnostic experiment graph $\mathcal G$ capturing experimental operations and their dependencies.
Based on $\mathcal G$ and the capabilities of available laboratory resources, automatable operations are lowered to validated device-specific code, while unsupported or intrinsically manual operations remain coordinated human-execution steps. This staged design separates scientific interpretation from hardware implementation and exposes intermediate artifacts for independent inspection and validation. Figure~\ref{fig:framework} illustrates the overall framework.

\subsection{Protocol Firewall: Diagnose Before Compiling}
\label{sec:protocol-firewall}

Protocols in realistic laboratory settings are not always execution-ready. Even when they describe a valid experimental goal, individual procedures may contain incorrect parameters, ambiguous operations, or inconsistencies across steps. 
\model{} places a protocol firewall before compilation to detect defects before they propagate into executable actions.

The firewall operates as an iterative \textit{diagnose-repair-verify} loop. From the original protocol $\mathcal P$, the screening agent identifies suspicious segments and proposes a minimally revised candidate $\mathcal P^{(t)}$.
Crucially, each repaired candidate is validated before acceptance, emphasizing comparison against the original protocol. The goal is to ensure that the correction resolves the diagnosed defect without causing unnecessary drift in unaffected experimental content, preserving valid operations, materials, targets, and dependencies while allowing revision of erroneous or inconsistent elements.
Formally, letting $\mathcal E^{(t)}$ denote the validation feedback at iteration $t$,
\begin{equation}
\begin{split}
\label{eq:diagnose_iteration}
     \mathcal P^{(t+1)} = \mathcal R\!\left( \mathcal P,\, \mathcal P^{(t)},\, \mathcal E^{(t)} \right), \qquad \mathcal E^{(t+1)} = \mathcal V\!\left( \mathcal P,\mathcal P^{(t+1)} \right),
\end{split}
\end{equation}
where \(\mathcal R\) denotes protocol diagnosis and repair, and \(\mathcal V\) 
validates whether the proposed repair resolves the diagnosed defect while preserving unaffected content and satisfying consistency and executability constraints.
Whenever validation fails, the original protocol, the revised candidate, and the validation feedback are returned to the repair agent for another revision. The loop terminates only when the candidate satisfies the required validation constraints; otherwise, revision proceeds for at most 12 iterations, after which the protocol is deemed unresolved. Importantly, scientific ambiguities that cannot be justified from available evidence are explicitly preserved for human clarification rather than resolved through unsupported assumptions.

Besides the repaired protocol, the firewall maintains a structured resolution trace for each accepted modification, diagnosed issue, supporting evidence, and applied correction. This preserves an explicit correspondence between the original and verified protocols, turning protocol repair into an auditable transformation rather than an unconstrained rewrite. The verified protocol $\widetilde{\mathcal P}$ then enters the subsequent representation stage.

\subsection{Perturbation-Guided Self-Evolving Diagnosis}
\label{sec:improvement}

To improve repair accuracy, \model{} introduces a perturbation-guided self-evolution mechanism converting controlled protocol perturbations in our benchmark into reusable troubleshooting knowledge.
The perturbations span multiple failure modes, as described in Section~\ref{sec:benchmark_design}.

Each self-evolution instance pairs a perturbed protocol with its verified reference, from which reusable troubleshooting experience is distilled without explicit error-category supervision. LabCompiler compares its diagnosis and repair with the reference and distills this experience into an external memory for subsequent retrieval. The memory is constructed from a protocol-disjoint training split of 100 instances per error category and frozen during held-out protocol evaluation.

\subsection{Device-agnostic Representation Abstraction}
\label{sec:semantic-compilation}

Natural-language protocols effectively support scientific communication but provide a poor execution interface. The same experimental operation may be expressed in substantially different forms, while implementation details such as repeated targets, implicit transfers, and compound operations are often linguistically underspecified. Directly generating device code from such descriptions would entangle scientific interpretation with low-level execution choices, making errors difficult to localize and the resulting workflow difficult to verify or reuse across platforms.

To provide a stable interface between protocol understanding and downstream execution, \model{} compiles the verified protocol $\widetilde{\mathcal P}$ into a device-agnostic experiment graph $\mathcal G$. Rather than committing immediately to device operations, the graph progressively resolves protocol semantics through two complementary representations:
\begin{equation}
     \widetilde{\mathcal P}
     \xrightarrow{\;\mathcal S\;}
     \mathcal G^{\mathrm{sem}}
     \xrightarrow{\;\mathcal N\;}
     \mathcal G^{\mathrm{can}},
 \end{equation}
 where $\mathcal S$ performs semantic interpretation and $\mathcal N$ denotes deterministic canonicalization.

 \vspace{-7pt}
 \paragraph{Semantic Experiment Graph.}
$\mathcal G^{\mathrm{sem}}$ captures the experimental semantics of the verified protocol $\widetilde{\mathcal P}$ without committing to any specific device implementation. It organizes the protocol into explicit operation-centric representations, where each operation specifies its participating materials, source and target objects, quantitative parameters, and experimental conditions, as illustrated in Figure~\ref{fig:framework}.
Beyond individual operations, $\mathcal G^{\mathrm{sem}}$ explicitly encodes control-flow and material-flow dependencies, transforming a sequential textual procedure into a structured graph of how experimental operations and material states evolve throughout the workflow.
Thus, $\mathcal G^{\mathrm{sem}}$ provides a device-independent representation preserving the scientific intent and procedural structure of the verified protocol while making its implicit experimental relationships explicit.

\vspace{-7pt}
\paragraph{Canonical Action Graph.}
The semantic graph is subsequently transformed into a canonical action graph $\mathcal G^{\mathrm{can}}$, expressing the experiment through a compact vocabulary of standardized laboratory actions.
This transformation reduces representational variability and aligns operations to a common granularity by standardizing parameters and units, decomposing compound procedures when necessary, and expanding repeated or multi-target operations into explicit action instances.
The current action space comprises 26 canonical action types, covering common laboratory operations such as \texttt{liquid\_transfer}, \texttt{plate\_reading} and \texttt{incubation}, and can be extended as additional experimental scenarios are incorporated. The complete action set appears in Appendix~\ref{app:canonical_actions}.
For example, add 100~$\mu$L of blue dye reagent to every occupied well is represented in $\mathcal G^{\mathrm{sem}}$ as a single operation over 17 targets, whereas $\mathcal G^{\mathrm{can}}$ expands it into 17 target-specific \texttt{liquid\_transfer} actions following the corresponding canonical semantics.
Together, $\mathcal G^{\mathrm{sem}}$ and $\mathcal G^{\mathrm{can}}$ form a progressive abstraction hierarchy: the former reconstructs and preserves protocol-level scientific semantics, while the latter projects them onto a uniform action space for downstream execution planning. Appendix~\ref{app:ir} provides the complete action set and representation details.

\subsection{Capability-Grounded Hybrid Execution}
\label{sec:execution}

Rather than assigning devices to canonical actions independently, \model{} first groups closely related actions into coherent execution units. Actions are grouped by their material flow, dependency structure, and whether they can be executed consecutively on the same device. For example, a sequence of liquid addition and mixing can be treated as a single execution unit when it operates on the same sample and can be executed within a compatible device.
For each execution unit $u_i$, \model{} selects an automated realization when a compatible device is available; otherwise, the unit is retained for human execution. Let $\mathcal D$ denote the available devices. In our current implementation, $\mathcal D$ comprises $45$ integrated devices, such as the Opentrons OT-2 and Flex~\citep{opentrons2026flex, opentrons2026ot2}, Hamilton STAR~\citep{hamilton2026star}, and Cytation plate readers.We define
\begin{equation}
     \pi(u_i)=
     \begin{cases}
         d,\quad d\in\mathcal D
         \text{ and } \Gamma(u_i,d)=1, & \text{if such } d \text{ exists},\\[2pt]
         \textsc{Human}, & \text{otherwise}.
     \end{cases}
 \end{equation}
Here, $\Gamma(u_i,d)=1$ indicates that device $d$ supports the required actions in the execution unit $u_i$. Thus, this local assignment allows a protocol to combine automated and manual operations without requiring the entire workflow to be executable on a single automation platform.

Once an execution unit is assigned to a device, its logical experimental objects are bound to concrete physical resources, such as labware, wells, and device locations. Device adapters then translate the corresponding canonical actions into platform-specific commands or code. This separation allows the same canonical experiment representation to be realized on different hardware without changing its underlying experimental semantics.
The generated device-specific code is first validated against the target device and resource constraints and, when a simulator is available, further verified through simulation. Only code passing these checks is accepted as a valid automated realization; otherwise, the automated compilation is considered unsuccessful.

\section{Protocol Correction Benchmark}
\label{sec:benchmark}

Existing benchmarks such as LAB-Bench~\citep{laurent2024lab} and BioProBench~\citep{liu2026bioprobench} evaluate protocol troubleshooting through multiple-choice questions or task-specific formulations such as error recognition and step ordering. In contrast, we evaluate protocol recovery: given an imperfect protocol without predefined corrective options, a system must identify and repair defects while preserving unaffected experimental information.
Our benchmark comprises two complementary subsets capturing controlled and naturally occurring protocol errors: a controlled perturbation subset, constructed by introducing structured defects into validated public protocols, and a real-world error subset, curated from experimental protocols written by iGEM competition teams~\citep{igem2023teamwikis} and annotated by domain experts for naturally occurring errors.

\begin{table*}[t]
\centering
\small
\caption{Taxonomy of controlled protocol perturbations.}
\label{tab:perturbation_taxonomy}

\renewcommand{\arraystretch}{1.1}
\setlength{\tabcolsep}{7pt}

\begin{tabularx}{\textwidth}{
     >{\raggedright\arraybackslash}p{0.20\textwidth}
     >{\raggedright\arraybackslash}X
     >{\raggedright\arraybackslash}p{0.27\textwidth}
 }
 \toprule
 \rowcolor{gray!10}
 \textbf{Perturbation}
 &
 \textbf{Construction}
 &
 \textbf{Illustrative example}
 \\
 \midrule

 \textbf{Parameter or reagent replacement}
 &
 Replace a valid experimental parameter or reagent with an incompatible
 alternative.
 &
 Incubate at $37^\circ\mathrm{C}$
 $\rightarrow$
 incubate at $4^\circ\mathrm{C}$.
 \\[2pt]

\addlinespace[3pt]
\textbf{Critical-step deletion}
&
Remove an operation required to establish the state assumed by a downstream
step.
&
Delete the quenching step before washing labeled cells.
\\[2pt]

\addlinespace[3pt]
\textbf{Cross-step inconsistency}
&
Modify a reference, quantity, or material such that two distant steps become inconsistent.
&
Prepare $N$ samples but later use reagents for $M$ samples.
 \\[2pt]

\addlinespace[3pt]
\textbf{Operation or parameter ambiguity}
&
Replace an explicit operation or quantitative specification with an expression insufficient to determine the intended procedure.
&
Add $500\,\mu\mathrm{L}$ of PBS
$\rightarrow$
add an appropriate amount of buffer.
\\

\bottomrule
\end{tabularx}
\vspace{-10pt}
\end{table*}

\subsection{Benchmark Construction}
\label{sec:benchmark_design}

\paragraph{Controlled Perturbation Subset.} 

We construct this subset from a quality-controlled collection of biological protocols from public sources, including SciRecipe~\citep{sun2026unleashing}. For each validated protocol, we introduce a targeted modification while leaving unrelated procedures unchanged, yielding perturbed instances with known error locations and reference corrections. This construction supports controlled evaluation of a model’s ability to identify and repair specific protocol defects.
We consider four perturbation families: parameter or reagent replacement, critical-step deletion, cross-step inconsistency, and operation or parameter ambiguity. Table~\ref{tab:perturbation_taxonomy} summarizes their construction. These perturbations are designed to alter the experimental meaning rather than the surface wording, while keeping the resulting protocol locally plausible. The held-out evaluation subset contains $2{,}000$ perturbed protocols, with $500$ instances per perturbation type.

\vspace{-8pt}
\paragraph{Real-World Error Subset.}
We further construct a real-world subset from experimental protocols written by participating teams in the iGEM competition. Unlike the controlled subset, these protocols contain errors that arise during actual protocol preparation rather than through predefined perturbation rules.
To ensure the validity of the data, we recruited domain experts to review the selected protocols, identify erroneous procedures, and provide scientifically valid corrections. Each erroneous protocol is paired with its expert-corrected reference version for evaluation. The resulting subset contains 259 instances and the detailed dataset statistics are provided in Appendix~\ref{app:benchmark}

\subsection{Quality Control and Evaluation Metric}
\label{sec:benchmark_evaluation}

To ensure quality and reliability, automated checks are applied to verify that each instance contains a single error and that the error is consistent with its assigned category. Biological domain experts then review 20\% of each error category to confirm that the labeled error constitutes a genuine experimental defect and that the corresponding reference protocol provides a valid correction.

The benchmark evaluates protocol recovery using two metrics. Classification accuracy measures whether the model correctly identifies the error type, while repair score measures the quality of the corrected protocol relative to the reference.
Because valid repairs may differ in wording, protocol repair is evaluated semantically rather than by exact string matching. A score of $2$ is assigned when the defect is fully resolved and the repair is scientifically consistent with the reference. A score of $1$ indicates a partially correct repair with a minor omission or inaccuracy, whereas a score of $0$ indicates that the defect remains unresolved or that the repair introduces a substantive error.

\section{Experiments}
\label{sec:experiments}

We evaluate \model{} along the three interfaces introduced by its staged design. Specifically, we test whether (i) the firewall can recover imperfect protocols before compilation, (ii) the device-agnostic representation preserves experimental semantics, and (iii) capability grounding reliably translates the representation into executable human-machine workflows. This decomposition allows failures to be attributed to individual stages rather than conflated into a single end-to-end score.
Additional component-wise ablations of the three stages are reported in Appendix~\ref{app:ablations}.

\subsection{Experimental Setup}
\label{sec:exp_setup}

We compare \model{} with a diverse set of frontier LLMs, including GPT-5.6 Sol~\citep{openai2026gpt56sol}, Claude Opus 5~\citep{anthropic2026claudeopus5}, DeepSeek-V4-Pro~\citep{deepseekai2026deepseekv4}, Gemini-3.1-Pro~\citep{googledeepmind2026gemini31pro}, Qwen3.8-Max~\citep{qwen2026qwen38max}, MiniMax-M3~\citep{minimax2026m3}, Kimi-K3~\citep{kimiteam2026kimik3}, and Codex(GPT-5.6 Sol) as a coding-agent baseline. Both \model{} and its self-evolving variant use DeepSeek-V4-Pro as the underlying model.
For categorical predictions, including error classification and execution routing, we report accuracy against the reference annotations. Metrics requiring semantic assessment are independently scored by two LLM judges, GPT-5.5 and Gemini-3.1-Pro. The scores are then averaged and normalized to a $0$-$100$ scale. Experts scored 30\% of model responses to assess agreement with LLM judges (Appendix~\ref{app:expert_judge}).

\begin{table*}[t]
\centering
\caption{Protocol correction on the controlled perturbation subset across four error categories. We report classification accuracy (Acc.) and repair score (Rep.). LabCompiler* denotes the self-evolved variant. Best and second-best results are shown in \textbf{bold} and \underline{underline}, respectively.}
\vspace{3pt}
\label{tab:perturbation_results_controlled}
\small
\setlength{\tabcolsep}{5pt}
\renewcommand{\arraystretch}{1.12}
\begin{tabular}{lcccccccc}
\toprule
\multicolumn{1}{c}{\multirow{2}{*}{\textbf{Method}}}
& \multicolumn{2}{c}{\makecell[c]{Critical-step\\deletion}}
& \multicolumn{2}{c}{\makecell[c]{Operation or parameter\\ambiguity}}
& \multicolumn{2}{c}{\makecell[c]{Parameter or reagent\\replacement}}
& \multicolumn{2}{c}{\makecell[c]{Cross-step\\inconsistency}}
\\
\cmidrule(lr){2-3}
\cmidrule(lr){4-5}
\cmidrule(lr){6-7}
\cmidrule(lr){8-9}
&
Acc.$\uparrow$ & Rep. $\uparrow$
& Acc.$\uparrow$ & Rep. $\uparrow$
& Acc.$\uparrow$ & Rep. $\uparrow$
& Acc.$\uparrow$ & Rep. $\uparrow$
\\
\midrule
MiniMax-M3       & 68.0 & 26.0 & 26.6 & 15.0 & 45.0 & 29.0 & 38.2 & 39.9 \\
Qwen3.8-Max      & 65.2 & 29.0 & 32.0 & 20.0 & 55.0 & 35.4 & 45.0 & 44.6 \\
DeepSeek-V4-Pro  & 69.8 & 37.3 & 37.0 & 23.7 & 49.0 & 35.0 & 47.4 & 47.7 \\
Gemini-3.1-Pro   & 68.0 & 36.4 & 39.8 & 22.0 & 52.0 & 32.0 & 51.4 & 45.9 \\
Kimi-K3          & 74.0 & 36.0 & 42.4 & 24.0 & 58.2 & 38.5 & 44.6 & 42.2 \\
GPT-5.6 Sol      & 79.6 & 39.0 & 48.2 & 26.8 & 72.6 & 52.6 & 48.4 & 55.3 \\
Claude Opus 5    & 78.4 & 38.9 & 47.4 & 25.3 & \underline{79.4} & \underline{61.2} & 54.4 & 56.2 \\
Codex+GPT-5.6 Sol             & \underline{81.8} & \underline{39.3} & 48.8 & 30.5 & 76.8 & 59.8 & 49.2 & 57.4 \\
\midrule
\textbf{LabCompiler}  & 80.0 & 39.1 & \underline{49.6} & \underline{30.6} & 78.8 & 54.0 & \underline{59.2} & \underline{64.0} \\
\textbf{LabCompiler*} & \textbf{93.8} & \textbf{44.4} & \textbf{64.2} & \textbf{35.3} & \textbf{85.6} & \textbf{63.9} & \textbf{66.8} & \textbf{72.4} \\
\bottomrule
\end{tabular}
\vspace{-5pt}
\end{table*}

\subsection{Protocol Correction}
\label{sec:exp_firewall}

\begin{figure}[t]
    \centering
    \includegraphics[width=\columnwidth]{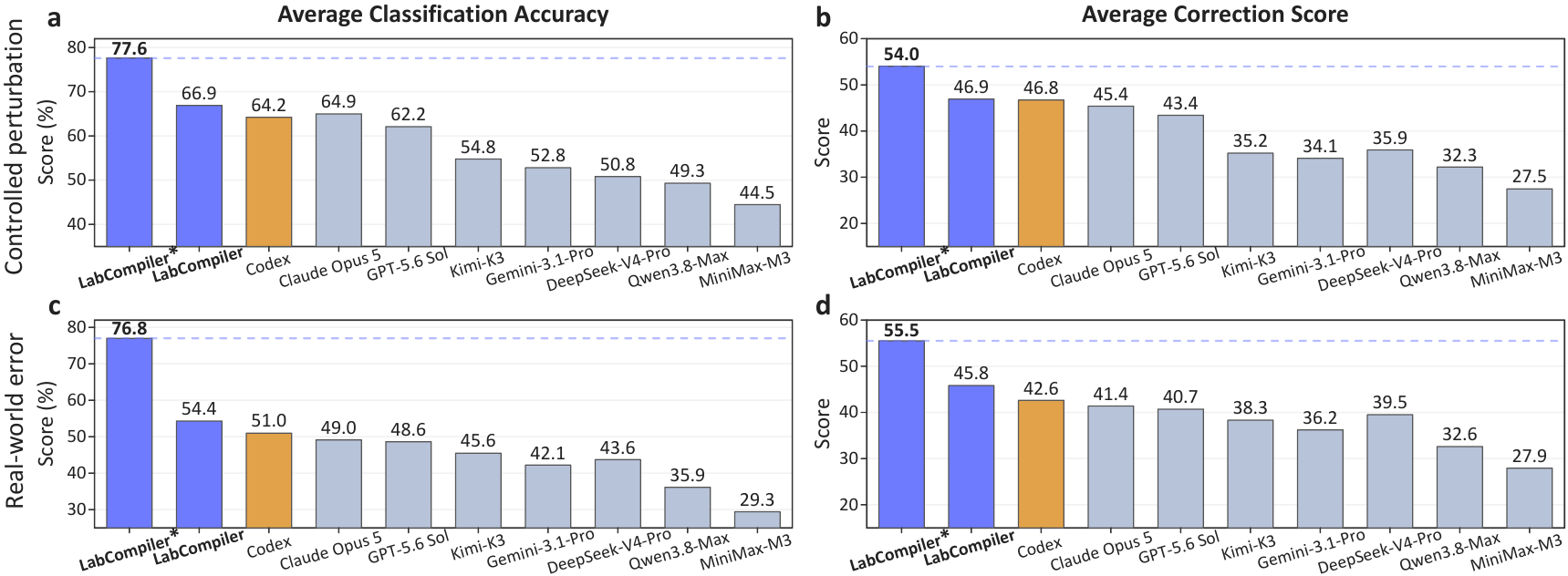}
    \vspace{-15pt}
    \caption{Average performance of protocol correction on the controlled perturbation subset (a,b) and the real-world error subset (c,d). LabCompiler* denotes the self-evolved variant. \model{} consistently improves both error classification and semantic repair over direct LLM baselines.}
    \label{fig:average_performance}
    \vspace{-5pt}
\end{figure}

\paragraph{Main results.}
Table~\ref{tab:perturbation_results_controlled} and Figure~\ref{fig:average_performance}(a,b) evaluate recovery from the four controlled perturbation families. Even without self-evolution, \model{} reaches an average classification accuracy of $66.9\%$ and a repair score of $46.9$, improving over the strongest direct LLM baseline by $2.0$ percentage points and $1.5$ score points, respectively. The comparison to its own DeepSeek-V4-Pro backbone is more pronounced: the proposed framework improves classification from $50.8\%$ to $66.9\%$ and repair from $35.9$ to $46.9$. The gain is particularly clear for failures that require reasoning across protocol context. \model{} obtains $59.2\%$ classification accuracy and a repair score of $64.0$ on cross-step inconsistencies, and reaches $49.6\%$ classification accuracy on operation or parameter ambiguity. 
Figure~\ref{fig:average_performance}(c,d) evaluates the same recovery task on the real-world error subset containing naturally occurring errors. LabCompiler achieves 54.4\% classification accuracy and a repair score of 45.8, compared with 43.6\% and 39.5 for its DeepSeek-V4-Pro backbone. With frozen self-evolved troubleshooting memory, LabCompiler* further improves to 76.8\% classification accuracy and a repair score of 55.5.
Notably, our method exhibits a larger advantage on naturally occurring errors than on controlled perturbations, indicating stronger robustness to heterogeneous defects encountered in practice. 
This advantage extends to LAB-Bench~\citep{laurent2024lab}, where \model{} consistently outperforms the baselines, as reported in Appendix~\ref{app:additional_results}.
By resolving these defects before compilation, the firewall reduces their propagation into downstream representation and execution.

\vspace{-7pt}
\paragraph{Effect of perturbation-guided self-evolution.}
We isolate the contribution of self-evolution while keeping the underlying language model fixed, as represented by two blue bars in Figure~\ref{fig:average_performance}(a,b). 
As the number of self-evolution examples per perturbation category increases from 0 to 50 and 100, average classification accuracy improves from 66.9\% to 72.5\% and 77.6\%, while the repair score increases from 46.9 to 50.1 and 54.0, respectively. At the 100-example-per-category setting, this yields absolute gains of 10.7 points in classification accuracy and 7.1 points in repair over LabCompiler without self-evolution.
Importantly, the gains are consistent across all four perturbation types, rather than being driven by a single category (Figure~\ref{fig:self_evolution}).
These results indicate that distilled troubleshooting experience improves both defect localization and the quality of correction.

\subsection{Analysis of Device-Agnostic Representation}
\label{sec:exp_ir}

A reliable intermediate representation should satisfy two properties: \emph{fidelity}, by preserving the experimental semantics of the source protocol, and \emph{stability}, by producing consistent execution structures for semantically equivalent protocol descriptions. We evaluate these two properties separately.

\begin{figure}[t]
    \centering
    \includegraphics[width=\columnwidth]{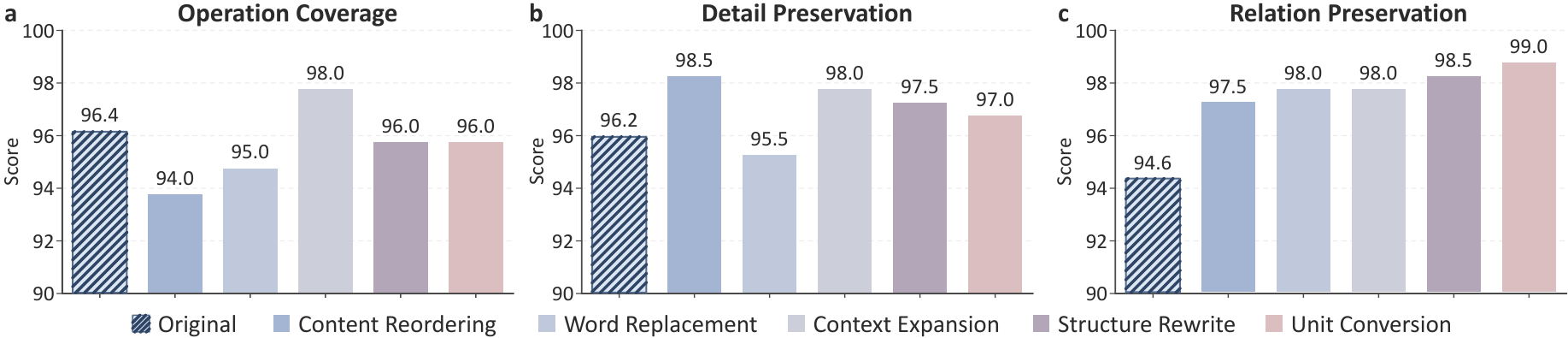}
    \vspace{-10pt}
    \caption{Fidelity and stability of the device-agnostic representation. \textit{Original} bars measure fidelity to the source protocol, while the other bars measure stability under five semantics-preserving rewrites, evaluated by (a) operation coverage, (b) detail preservation, and (c) relation preservation.}
    \label{fig:protocoltoIR}
\vspace{-15pt}
\end{figure}

\vspace{-7pt}
\paragraph{Representation fidelity.}
We first evaluate whether the generated IR faithfully preserves the source protocol. Two independent judges assess each IR along three dimensions: \emph{operation coverage}, which measures whether the required experimental operations are correctly represented without unsupported additions; 
\emph{detail preservation}, which measures retention of execution-relevant materials, parameters, and conditions; 
and \emph{relation preservation}, which measures whether dependencies and relationships among operations are correctly maintained. 
As shown by the original bars in Figure~\ref{fig:protocoltoIR}, \model{} achieves scores of $96.4$, $96.2$, and $94.6$ on the three dimensions, respectively. The consistently high scores show that the device-agnostic abstraction preserves both the experimental operations and the execution-relevant details of the source protocol. Relation preservation is comparatively more challenging, indicating that errors are more likely to arise when reconstructing dependencies across operations than when preserving individual operations or parameters.

\vspace{-7pt}
\paragraph{Representation stability.}
We next examine whether semantically equivalent descriptions lead to consistent representations. For each protocol, we generate ten rewrites spanning five meaning-preserving transformations: content reordering, word replacement, context expansion, structure rewrite, unit conversion. We then compare the resulting IRs with the representation generated from the original protocol.
As shown in Figure~\ref{fig:protocoltoIR}, \model{} remains highly consistent across all five transformations, with average scores of $95.8$ for operation coverage, $97.3$ for detail preservation, and $98.2$ for relation preservation.
These results show that the inferred execution structure is largely insensitive to surface-level variations in protocol description, allowing semantically equivalent protocols to be mapped to consistent device-agnostic representations.

\begin{figure}[t]
    \centering
    \includegraphics[width=\columnwidth]{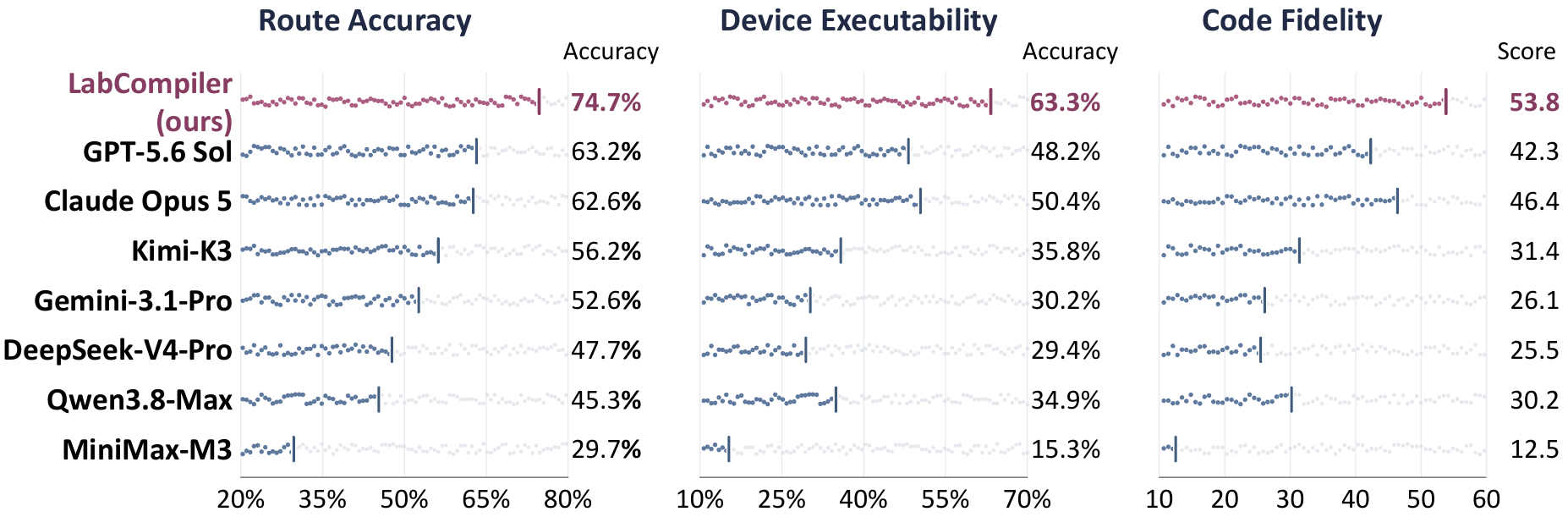}
    \vspace{-10pt}
    \caption{Execution performance across route accuracy, device executability, and code fidelity. \model{} consistently outperforms direct LLM baselines across all three metrics.}
    \label{fig:protocotocode}
\vspace{-10pt}    
\end{figure}

\subsection{Evaluation of Hybrid Execution}
\label{sec:exp_execution}
The final evaluation examines whether the canonical representation can be grounded into an executable human-machine workflow while preserving the intended experimental semantics. We evaluate three complementary aspects of this process: \emph{route accuracy}, \emph{device executability}, and \emph{code fidelity}. 
Specifically, route accuracy measures whether each execution unit is assigned to the expert-annotated executor under the available device capabilities, $i.e.$, a compatible device or a human operator.
Device executability is evaluated on units annotated for automated execution.
Generated code is considered successful only if it satisfies the target-device and resource constraints defined by the corresponding device interface and, when an official simulator or device validation pathway is available, passes the corresponding validation procedure.
Finally, code fidelity evaluates whether executable code still realizes the intended experiment. 
Following the scoring scheme described above, two LLM judges assess each generated program along three dimensions on a $0$-$5$ scale: \emph{action correctness}, \emph{experimental constraint preservation}, and \emph{logic preservation}. The three dimensions are averaged and normalized to a $0$-$100$ scale.

As shown in Figure~\ref{fig:protocotocode}, \model{} achieves $74.7\%$ route accuracy, $63.3\%$ device executability, and a code-fidelity score of $53.8$, outperforming the strongest direct baseline on each metric by $11.5$, $12.9$, and $7.4$ points, respectively. Notably, stronger routing does not necessarily lead to better downstream execution: GPT-5.6 Sol slightly exceeds Claude Opus 5 in route accuracy ($63.2\%$ vs.\ $62.6\%$), but underperforms it in both device executability ($48.2\%$ vs.\ $50.4\%$) and code fidelity ($42.3$ vs.\ $46.4$). This suggests that selecting an appropriate executor is only one part of successful execution, and that preserving protocol semantics during device-specific realization remains a separate challenge. 
With DeepSeek-V4-Pro fixed as the backbone, \model{} improves route accuracy, device executability, and code fidelity by $27.0$, $33.9$, and $28.3$ points over direct inference, respectively. The consistent gains suggest that the proposed framework, particularly its structured representation, provides a more reliable bridge from protocol semantics to device-specific execution. These metrics assess execution grounding rather than wet-lab outcomes.
\section{Conclusions and Limitations}
\label{sec:conclusion}

We introduced \model{}, a staged framework for translating imperfect natural-language protocols into verifiable human-machine laboratory workflows. By separating protocol correction, device-agnostic representation, and capability-grounded execution, \model{} improves protocol recovery, preserves experimental semantics, and supports capability-grounded execution planning across heterogeneous laboratory settings. Experiments on both controlled and naturally occurring protocol errors further show that perturbation-guided self-evolution consistently strengthens correction performance. 
Our current implementation covers a finite action space and set of device backends. Future work will expand this coverage to broader laboratory settings and further validate the framework through real-world wet-lab experiments.

\bibliography{iclr2027_conference}
\bibliographystyle{iclr2027_conference}

\newpage
\appendix

\startcontents[si]

\section*{Contents}
\printcontents[si]{}{1}[2]{}

\clearpage


\setcounter{table}{0}
\setcounter{figure}{0}

\renewcommand{\thetable}{\Alph{section}.\arabic{table}}
\renewcommand{\thefigure}{\Alph{section}.\arabic{figure}}


\section{Comparison with Closely Related Systems}

Recent studies have explored different components of laboratory automation,
including structured protocol representations, protocol reasoning and
verification, robotic execution, and agentic scientific workflow generation.
Table~\ref{tab:related_systems} summarizes capabilities explicitly described
in closely related systems that are relevant to reliable protocol-to-execution
translation.

We use three categories to describe each capability. ``Supported'' indicates
that the capability is explicitly described or implemented in the original
publication. ``Partial'' indicates that the work contains a related mechanism
but differs in scope, objective, or level of integration. ``Not explicitly
reported'' indicates that the corresponding capability is not described in the original publication and does not imply that the system cannot support it.
The table is intended as a qualitative summary of reported system components
rather than a unified empirical benchmark across systems.

\begin{table}[!htbp]
\centering
\small

\setlength{\tabcolsep}{1.5pt}
\renewcommand{\arraystretch}{1.12}
\vspace{-10pt}
\caption{
Qualitative summary of explicitly reported capabilities in closely related
laboratory automation systems. Hierarchical representation refers to systems
that expose multiple abstraction levels or explicit transformation stages
between protocol descriptions and execution representations. Human hybrid
workflow support refers to explicit coordination between automated and
human-performed operations rather than the general possibility of manual
intervention.
}
\label{tab:related_systems}
\vspace{5pt}
\begin{tabular}{
@{}
>{\raggedright\arraybackslash}p{0.16\textwidth}
>{\centering\arraybackslash}p{0.11\textwidth}
>{\centering\arraybackslash}p{0.11\textwidth}
>{\centering\arraybackslash}p{0.10\textwidth}
>{\centering\arraybackslash}p{0.11\textwidth}
>{\centering\arraybackslash}p{0.11\textwidth}
>{\centering\arraybackslash}p{0.11\textwidth}
>{\centering\arraybackslash}p{0.12\textwidth}
@{}
}

\toprule

\textbf{Method}
&
\textbf{Protocol defect repair}
&
\textbf{Explicit protocol verification}
&
\textbf{Device-agnostic IR}
&
\textbf{Hierarchical representation}
&
\textbf{Device capability grounding}
&
\textbf{Human hybrid workflow}
&
\textbf{Auditable intermediate artifacts}
\\

\midrule

\makecell[l]{XDL\\ \citep{mehr2020universal}}
&
Partial
&
Supported
&
Supported
&
Partial
&
Partial
&
Not explicitly reported
&
Supported
\\

\addlinespace[2pt]

\makecell[l]{LabOP\\ \citep{bartley2023building}}
&
Not explicitly reported
&
Supported
&
Supported
&
Partial
&
Partial
&
Supported
&
Supported
\\

\addlinespace[2pt]

\makecell[l]{BPL\\ \citep{song2026towards}}
&
Partial
&
Supported
&
Supported
&
Partial
&
Supported
&
Supported
&
Partial
\\

\addlinespace[2pt]

\makecell[l]{BioProAgent\\ \citep{liu2026bioproagent}}
&
Partial
&
Supported
&
Partial
&
Partial
&
Supported
&
Partial
&
Partial
\\

\addlinespace[2pt]

\makecell[l]{ProtoAct\\ \citep{liu2026protoact}}
&
Partial
&
Partial
&
Partial
&
Partial
&
Supported
&
Partial
&
Partial
\\

\addlinespace[2pt]

\makecell[l]{ProtoPilot\\ \citep{jiang2026self}}
&
Supported
&
Supported
&
Partial
&
Partial
&
Supported
&
Partial
&
Supported
\\

\addlinespace[3pt]

\textbf{LabCompiler}
&
\textbf{Supported}
&
\textbf{Supported}
&
\textbf{Supported}
&
\textbf{Supported}
&
\textbf{Supported}
&
\textbf{Supported}
&
\textbf{Supported}
\\

\bottomrule

\end{tabular}
\end{table}

The comparison emphasizes a distinction that is central to LabCompiler: laboratory workflows may combine automated operations with human observation, judgment, or manipulation when available devices cannot fully execute a protocol. LabCompiler retains these operations as explicit human-coordinated steps while preserving intermediate artifacts for inspection.

\section{Protocol Firewall and Self-Evolving Diagnosis}

\subsection{SCREENING CRITERIA AND DECISION RULES}

The protocol firewall applies an iterative diagnose–repair–screen procedure before semantic compilation. In this work, a “verified” protocol is an internal status assigned after the protocol passes the firewall’s screening criteria. It indicates that the diagnosed defect has been addressed, unaffected content has been preserved, and the applicable consistency and ambiguity-handling checks have passed. It does not constitute a formal proof of scientific correctness or physical laboratory validation.
The firewall evaluates four major aspects:

\begin{itemize}[leftmargin=*]
    \item \textbf{Report validity}: the screening output must satisfy the
    required protocol report format and contain the necessary effective
    protocol artifact.
    
    \item \textbf{Content preservation}: explicit experimental information in
    the original protocol, including operations, materials, preparation
    requirements, and relevant constraints, must be preserved unless a
    justified correction is required.
    
    \item \textbf{Protocol consistency}: the repaired protocol must maintain
    consistency among operations, materials, parameters, and dependencies.
    
    \item \textbf{Ambiguity handling}: remaining ambiguities are classified
    according to whether they prevent construction of a valid executable
    representation.
\end{itemize}

The verification process distinguishes deterministic checks from
scientific-context judgments. Deterministic checks are enforced by the
screening system and cannot be overridden by the model's self-reported
status. Scientific-context decisions determine whether missing information can
be reasonably resolved or should remain explicitly represented as ambiguity. Table~\ref{tab:firewall_checks} summarizes the screening criteria.

\begin{table}[t]
\centering
\caption{
Summary of protocol firewall screening criteria.
}
\label{tab:firewall_checks}
\begin{tabular}{p{0.23\linewidth}p{0.32\linewidth}p{0.32\linewidth}}
\toprule
Category & Check & Failure handling
\\
\midrule

Report structure &
Required screening fields and protocol artifact consistency &
The screening report is rejected and returned for revision
\\

Experimental content preservation &
Preservation of explicit operations, materials, preparation commitments,
and protocol constraints &
The candidate protocol is rejected if valid content is removed or altered
without justification
\\

Consistency checking &
Consistency among steps, materials, parameters, and dependencies &
Feedback is returned for another repair attempt
\\

Ambiguity classification &
Classification of remaining ambiguities into blocking or deferred cases &
Blocking ambiguity leads to unresolved status; deferred ambiguity may be
carried forward
\\

Evidence-supported completion &
Completion of missing information only when supported by available evidence
or justified inference &
Unsupported assumptions remain unresolved rather than being inserted
\\

\bottomrule
\end{tabular}
\end{table}

The firewall uses four execution states:

\begin{itemize}[leftmargin=*]
    \item \textbf{Accept}: all applicable screening constraints are satisfied
    and an approved effective protocol is generated.
    \item \textbf{Revise}: the submitted screening result fails a screening
    constraint and is returned for another repair attempt.
    \item \textbf{Unresolved}: protocols marked unresolved are excluded from downstream automated execution.
    \item \textbf{Clarification required}: remaining ambiguity is preserved
    for human clarification when it does not prevent continued execution
    planning.
\end{itemize}

Clarification required indicates that additional human input is needed before
the corresponding unresolved interpretation can be finalized. Such cases are
kept explicit in the workflow artifacts and are not treated as automatically
resolved scientific decisions. Remaining ambiguities that do not block representation may be retained as
explicit clarification items for human review. Such items are preserved in the
workflow artifacts rather than silently resolved by unsupported assumptions. In contrast, blocking
ambiguities prevent generation of an approved effective protocol and stop
downstream compilation.

The diagnose--repair--verify loop follows the iterative procedure described in
the main text. The revision process continues for at most 12 iterations. If
the protocol does not satisfy the screening requirements within this budget,
the protocol is marked unresolved.

\subsection{Minimal Repair and Resolution Traces}

The repair process follows a minimal modification principle. The objective is
to resolve diagnosed defects while preserving unaffected experimental
information. The screening module avoids introducing additional operations,
parameters, or assumptions that are not supported by the protocol context.
For each accepted modification, the firewall records a structured resolution
trace. The trace provides correspondence between the original protocol and
the accepted effective protocol, enabling inspection of how each correction
was obtained.

A resolution record contains the following information:
Resolution traces are also used to distinguish protocol-level clarification
from execution-level human involvement. Scientific ambiguities that cannot be
resolved from available evidence are preserved as remaining ambiguities rather
than being replaced by unsupported assumptions.
Human clarification and human execution represent different mechanisms.
Human clarification occurs during protocol screening when scientific
interpretation remains unresolved. Human execution occurs after representation
and routing when an operation cannot be automatically executed using available
laboratory capabilities.

\subsection{Memory Construction and Frozen Evaluation}

LabCompiler uses perturbation-guided self-evolution to construct an external troubleshooting memory from corrected protocol-recovery examples. Each self-evolution example consists of a perturbed protocol and its reference correction. LabCompiler first performs protocol diagnosis and repair on the perturbed protocol, compares the resulting diagnosis and correction against the reference correction, and then distills reusable troubleshooting experience into memory. The resulting memory is intended to capture transferable diagnostic and repair patterns rather than to store or reproduce individual protocol instances.
Memory entries summarize reusable information such as characteristic error patterns, contextual conditions under which a pattern may apply, relevant screening considerations, and correction guidance. The underlying language-model parameters are not updated during this process; self-evolution only updates the external troubleshooting memory used by the protocol firewall.

For controlled analysis, we vary the amount of protocol-recovery experience used for memory construction. Let $K$ denote the number of self-evolution examples sampled from each perturbation category. We evaluate $K \in \{0, 50, 100\}$. Because the controlled benchmark contains four perturbation categories, these settings correspond to 0, 200, and 400 memory-construction examples in total, respectively. The per-category allocation is used only to construct balanced self-evolution pools. Importantly, perturbation-category labels are not provided to LabCompiler during memory construction; the system learns troubleshooting patterns from the protocol--reference pairs without explicit error-category supervision.

The full self-evolution split contains 100 examples from each perturbation category, for a total of 400 examples. These examples are separated from the 2,000 held-out controlled evaluation instances described in Appendix~E. The 50-example-per-category condition uses a subset of 200 self-evolution examples, whereas the 100-example-per-category condition uses the complete 400-example self-evolution split.
For each evaluation condition, troubleshooting memory is constructed before evaluation and then frozen. Held-out evaluation protocols, their reference corrections, and their evaluation annotations are not used to update the memory during testing. Consequently, differences among the $K=0$, $K=50$, and $K=100$ conditions reflect different amounts of previously accumulated troubleshooting experience while keeping the evaluation set and the underlying language model fixed.

\section{Device-Agnostic Intermediate Representations}
\label{app:ir}

\subsection{Semantic Experiment Graph Schema}
\label{app:semantic_ir}

The Semantic Experiment Graph, $G_{\mathrm{sem}}$, represents the
experimental semantics of the approved effective protocol before
device-specific realization. An operation is represented as a logical
node, while predecessor relationships are represented through explicit
dependencies. Participating materials, samples, containers, and
collections are represented as logical objects rather than physical
device resources. The serialized representation contains an \texttt{operations} array
and, when applicable, an \texttt{analysis\_annotations} array.
Table~\ref{tab:semantic_ir_schema} summarizes the operation-level
schema.

\begin{table}[t]
\centering
\small
\setlength{\tabcolsep}{4pt}
\renewcommand{\arraystretch}{1.15}
\caption{Operation-level fields of the Semantic Experiment Graph.}
\label{tab:semantic_ir_schema}
\begin{tabular}{
@{}
>{\raggedright\arraybackslash}p{0.23\linewidth}
>{\raggedright\arraybackslash}p{0.16\linewidth}
>{\raggedright\arraybackslash}p{0.54\linewidth}
@{}
}
\toprule
Field & Type & Description \\
\midrule

\texttt{operation\_id}
&
String
&
Unique identifier of the semantic operation.
\\

\texttt{operation\_type}
&
String
&
Semantic operation type before canonical action normalization.
\\

\texttt{source\_text}
&
String
&
Exact excerpt from the approved effective protocol supporting
the operation.
\\

\texttt{objects}
&
Array
&
Logical materials, samples, containers, or collections participating
in the operation, together with their semantic roles.
\\

\texttt{parameters}
&
Object
&
Quantitative parameters, experimental conditions, and other
operation-level values explicitly represented by the protocol.
\\

\texttt{dependencies}
&
Array
&
Identifiers of predecessor operations.
\\

\texttt{operation\_details}
&
Object
&
Additional semantic information that does not belong to the
dependency structure or primary parameter set.
\\

\bottomrule
\end{tabular}
\end{table}

\paragraph{Logical objects and material flow.}
Objects may contain a stable logical identifier, a human-readable
name, a semantic role, an object type, and attributes that are
explicitly established by the protocol. Examples include a sample,
solution, container, plate, or collection of wells.
Object identity is kept separate from object state. The same logical
sample can therefore be referenced by successive operations without
creating a new identity solely because its state has changed.
Material flow is represented jointly by logical object reuse and
operation dependencies.
The semantic representation does not bind these objects to a physical
well, deck slot, pipette, or device instance. Such bindings are
introduced only during execution grounding.

\paragraph{Control and dependencies.}
$G_{\mathrm{sem}}$ represents procedural ordering and prerequisite relations through explicit dependencies between operations.
Control-related information, including conditions, repetitions, and qualitative endpoints, is retained as semantic attributes of the corresponding operations when it is part of the approved protocol.
When an operation depends on independently prepared branches, its
dependencies preserve the relevant predecessor operations rather than
collapsing the branches into a single implicit ordering.

\vspace{-8pt}
\paragraph{Source correspondence.}
Each semantic operation retains a \texttt{source\_text} field that
points to the corresponding text in the approved effective protocol.
The correspondence from the original raw protocol to the approved
effective protocol is maintained separately through the resolution
traces described in Appendix~B.

\paragraph{Missing values.}
Empty object, parameter, and dependency collections are represented as
\texttt{[]}, \texttt{\{\}}, and \texttt{[]}, respectively.
An unknown numerical value is represented by omitting the corresponding
parameter key rather than inserting a numerical default.

When a value has been explicitly declared as a runtime quantity, it may
be represented as

\begin{verbatim}
{"runtime_input": "<declared_name>"}
\end{verbatim}

Derived values may retain a formula together with the runtime inputs
referenced by that formula. Qualitative protocol descriptions remain
qualitative unless a numerical value is explicitly established by the
protocol or by the preceding screening stage.

\paragraph{Analysis annotations.}
Protocol statements that describe analysis rather than material
manipulation can be stored separately in
\texttt{analysis\_annotations}. An annotation records its identifier,
type, source text, related operation identifiers, and optional details.
For example, constructing a standard curve after a plate measurement
can be retained as an analysis annotation rather than being converted
into an execution action.

\paragraph{Illustrative serialization.}
For the protocol fragment

\begin{quote}
Transfer 200 uL of sample\_alpha to destination\_alpha.\\
Measure absorbance of each assay reaction at 450 nm.\\
Build a standard curve and calculate each sample concentration.
\end{quote}

a simplified semantic representation is:

\begin{lstlisting}[
    style=json,
    caption={Example JSON output of the structured representation of laboratory operations and post-measurement analysis.},
    label={lst:example-json}
]
{
  "operations": [
    {
      "operation_id": "op-001",
      "operation_type": "liquid_transfer",
      "source_text": "Transfer 200 uL of sample_alpha to destination_alpha.",
      "objects": [
        {
          "object_id": "sample_alpha",
          "name": "sample_alpha",
          "role": "source"
        },
        {
          "object_id": "destination_alpha",
          "name": "destination_alpha",
          "role": "destination"
        }
      ],
      "parameters": {
        "volume": "200 uL"
      },
      "dependencies": [],
      "operation_details": {}
    },
    {
      "operation_id": "op-measure",
      "operation_type": "measure_absorbance",
      "source_text": "Measure absorbance of each assay reaction at 450 nm.",
      "objects": [
        {
          "name": "assay reactions",
          "role": "measurement targets"
        }
      ],
      "parameters": {
        "wavelength": "450 nm"
      },
      "dependencies": ["op-001"],
      "operation_details": {}
    }
  ],
  "analysis_annotations": [
    {
      "annotation_id": "ann-standard-curve",
      "annotation_type": "post_measurement_analysis",
      "source_text": "Build a standard curve and calculate each sample concentration.",
      "related_operation_ids": ["op-measure"]
    }
  ]
}
\end{lstlisting}

\subsection{Canonical Action Graph Schema and the 26 Canonical Actions}
\label{app:canonical_actions}

The Canonical Action Graph, $G_{\mathrm{can}}$, expresses the
experiment using a standardized vocabulary of 26 laboratory action
types. The representation remains device-agnostic: canonical actions
specify what operation is required without selecting a physical
executor or binding the operation to device-specific resources.
Each normalized operation preserves its correspondence to
$G_{\mathrm{sem}}$ and contains the fields summarized in
Table~\ref{tab:canonical_ir_schema}.

\begin{table}[t]
\centering
\small
\setlength{\tabcolsep}{4pt}
\renewcommand{\arraystretch}{1.15}
\caption{Operation-level fields of the Canonical Action Graph.}
\label{tab:canonical_ir_schema}
\begin{tabular}{
@{}
>{\raggedright\arraybackslash}p{0.37\linewidth}
>{\raggedright\arraybackslash}p{0.57\linewidth}
@{}
}
\toprule
Field & Description \\
\midrule

\texttt{normalized\_operation\_id}
&
Unique identifier of the normalized operation.
\\

\texttt{canonical\_action}
&
One of the 26 canonical action identifiers listed in
Table~\ref{tab:canonical-actions}.
\\

\texttt{sequence\_index}
&
Position of the operation in the normalized operation sequence.
\\

\texttt{derived\_from\_semantic
\_operation\_ids}
&
Semantic operations from which the canonical operation is derived.
\\

\texttt{source\_texts}
&
Source excerpts corresponding to the contributing semantic operations.
\\

\texttt{objects}
&
Mapping between action-specific logical roles and logical objects.
\\

\texttt{parameters}
&
Action-specific experimental parameters.
\\

\texttt{dependencies}
&
Predecessor canonical operations.
\\

\texttt{operation\_details}
&
Additional action-specific semantic details.
\\

\texttt{control}
&
Condition, repetition, endpoint, or constraint metadata when present.
\\

\texttt{normalization\_trace}
&
Provenance of the semantic-to-canonical transformation.
\\

\bottomrule
\end{tabular}
\end{table}

\paragraph{Object roles.}
Unlike the object array in $G_{\mathrm{sem}}$, the
\texttt{objects} field in $G_{\mathrm{can}}$ is organized using
action-specific logical roles. For example, a liquid-transfer
operation may contain \texttt{source} and \texttt{destination},
whereas an incubation operation primarily acts on a \texttt{sample}.
These roles remain logical. They do not encode a physical deck
location, well allocation, or selected laboratory instrument.

\paragraph{Quantities and units.}
An explicit numerical quantity may be represented as follows, such that both the normalized quantity and its original textual form are retained:

\begin{lstlisting}[style=json]
{
  "value": 200,
  "unit": "uL",
  "source_value": "200 uL"
}
\end{lstlisting}

Canonical unit labels include \texttt{uL} and \texttt{mL} for volume;
\texttt{celsius} for temperature; \texttt{second},
\texttt{minute}, \texttt{hour}, and \texttt{day} for time;
\texttt{rpm} for rotational speed; \texttt{g} for relative
centrifugal force; and \texttt{nm} for wavelength. Missing values are not introduced solely to populate the schema.
Device-specific capability ranges and resource requirements are applied during execution grounding and are described in Appendix~D.


\begingroup
\small

\setlength{\tabcolsep}{4pt}
\renewcommand{\arraystretch}{1.12}
\setlength{\LTleft}{0pt}
\setlength{\LTright}{0pt}
\setlength{\LTcapwidth}{\textwidth}

\begin{longtable}{
    @{}
    >{\raggedright\arraybackslash}p{0.24\textwidth}
    >{\raggedright\arraybackslash}p{0.42\textwidth}
    >{\raggedright\arraybackslash}p{0.29\textwidth}
    @{}
}

\caption{Complete vocabulary of the 26 canonical actions, including their experimental semantics, logical roles, and key semantic restrictions.}
\label{tab:canonical-actions}
\\

\toprule
\rowcolor{black!6}
\textbf{Canonical action}
& \textbf{Semantics and logical roles}
& \textbf{Key semantic restriction} \\
\midrule
\endfirsthead

\multicolumn{3}{@{}l}{\textit{Table C.5 continued from previous page}} \\[2pt]

\toprule
\rowcolor{black!6}
\textbf{Canonical action}
& \textbf{Semantics and logical roles}
& \textbf{Key semantic restriction} \\
\midrule
\endhead

\midrule
\multicolumn{3}{r@{}}{\textit{Continued on next page}} \\
\endfoot

\bottomrule
\endlastfoot

\path{liquid_transfer}
&
Move liquid from one logical source to one logical destination.
&
Single-source, single-destination liquid movement; it does not represent mixing or liquid disposal.
\\[3pt]

\path{liquid_distribution}
&
Distribute liquid from one source to multiple logical destinations.
&
Used when one source supplies a logical destination collection.
\\[3pt]

\path{liquid_consolidation}
&
Collect liquid from multiple logical sources into one destination.
&
Distinguished from a single-source liquid transfer.
\\[3pt]

\path{liquid_removal}
&
Remove liquid from a source to a discard destination; a protected material may remain in the source.
&
Represents removal rather than productive transfer of the removed liquid.
\\[3pt]

\path{non_liquid_sample_transfer}
&
Move a discrete non-liquid sample to a logical destination.
&
Used for non-liquid sample movement rather than liquid transfer.
\\[3pt]

\path{solid_combination}
&
Combine a solid component and a liquid component into a resulting product.
&
Material phase must be supported by the interpreted protocol.
\\[3pt]

\path{non_liquid_material_removal}
&
Remove a non-liquid material from a source while retaining the relevant remaining material.
&
Distinguished from liquid removal and whole-sample transfer.
\\[3pt]

\path{pipette_mixing}
&
Mix a sample through repeated aspiration and dispensing.
&
Preserves explicit pipette-mixing semantics.
\\[3pt]

\path{vortex_mixing}
&
Mix a sample by vortexing.
&
Vortexing is retained as a distinct mixing mechanism.
\\[3pt]

\path{orbital_shaking}
&
Apply orbital shaking to a sample.
&
Preserves orbital-shaking semantics.
\\[3pt]

\path{mechanical_stirring}
&
Mechanically stir a sample.
&
Used when a mechanical stirring mechanism is specified.
\\[3pt]

\path{manual_inversion}
&
Mix a sample by repeatedly inverting its container.
&
Preserves inversion rather than replacing it with another mixing mechanism.
\\[3pt]

\path{incubation}
&
Maintain a sample under stated experimental conditions for a duration or until a stated endpoint.
&
Used when the operation carries incubation semantics.
\\[3pt]

\path{plate_reading}
&
Measure a sample using a stated plate-reading modality.
&
Measurement modality and wavelengths are preserved without selecting a physical reader.
\\[3pt]

\path{waiting}
&
Represent a protocol-defined delay or waiting period.
&
Used when the operation does not carry incubation semantics.
\\[3pt]

\path{temperature_control}
&
Apply or maintain a non-cyclic temperature condition.
&
Distinguished from a repeated multi-step temperature program.
\\[3pt]

\path{thermal_cycling}
&
Apply a repeated multi-step temperature program.
&
Represents an explicitly structured thermal-cycling program.
\\[3pt]

\path{magnetic_separation}
&
Separate magnetic material from or within a sample using a magnetic field.
&
Requires magnetic-separation semantics.
\\[3pt]

\path{plate_tilting}
&
Tilt a plate to a stated orientation.
&
Tilting does not itself imply shaking or inversion.
\\[3pt]

\path{centrifugation}
&
Apply centrifugation to a sample.
&
Rotational speed and relative centrifugal force remain distinct quantities.
\\[3pt]

\path{plate_washing}
&
Represent a protocol-level washing operation over a plate or well collection.
&
Retains high-level plate-washing semantics.
\\[3pt]

\path{plate_sealing}
&
Seal a sample plate.
&
Sealing conditions are retained when specified.
\\[3pt]

\path{plate_unsealing}
&
Remove a seal from a sample plate.
&
Distinguished from applying a plate seal.
\\[3pt]

\path{colony_picking}
&
Pick specified colonies from a sample plate.
&
Requires colony-picking semantics in the interpreted protocol.
\\[3pt]

\path{controlled_environment}
&
Operate under explicitly stated environmental conditions.
&
Retains explicitly stated environmental conditions without selecting a physical environment controller.
\\[3pt]

\path{reagent_dispensing}
&
Dispense reagent using dedicated dispensing semantics.
&
Distinguished from ordinary liquid transfer or distribution.
\\

\end{longtable}

\endgroup

\subsection{Deterministic Canonicalization Rules}
\label{app:canonicalization}

As defined in the main text, the approved effective protocol is first
mapped to $G_{\mathrm{sem}}$ through semantic interpretation and then
normalized into $G_{\mathrm{can}}$:

\begin{equation}
\widetilde{P}
\xrightarrow{\mathcal{S}}
G_{\mathrm{sem}}
\xrightarrow{\mathcal{N}}
G_{\mathrm{can}} .
\end{equation}

Here, $\mathcal{S}$ resolves protocol-level semantics, whereas
$\mathcal{N}$ standardizes the resulting representation into the
canonical action space. The main deterministic transformations are
described below.

\paragraph{Parameter and unit normalization.}
Equivalent textual quantity forms are mapped to a common
representation. For example, \texttt{"200 uL"} is normalized as

\begin{lstlisting}[style=json]
{
  "value": 200,
  "unit": "uL",
  "source_value": "200 uL"
}
\end{lstlisting}

Unit aliases are normalized without changing the underlying stated
quantity. For example, microliter variants are represented as
\texttt{uL}, Celsius temperature as \texttt{celsius}, and common time
abbreviations are mapped to their canonical time-unit labels.
Normalization does not introduce a numerical value solely because a
parameter is absent. Qualitative experimental conditions remain
qualitative unless a quantitative value has already been established
by the protocol interpretation.

\paragraph{Canonical operation granularity.}
$G_{\mathrm{can}}$ projects semantically equivalent operations onto a
common action vocabulary. Compound operations may therefore be
decomposed into multiple canonical operations when required to expose
the executable action structure. Conversely, multiple semantic
fragments that jointly describe one executable action can retain their
combined lineage in the normalized representation.
The normalization trace records the correspondence using four mapping
modes: \texttt{direct}, \texttt{decomposed}, \texttt{merged}, and
\texttt{decomposed\_and\_merged},
together with the contributing semantic operation identifiers and
source text.

\paragraph{Repeated and multi-target operations.}
Repeated or multi-target procedures are made explicit in
$G_{\mathrm{can}}$. In particular, the example described in the main
text,

\begin{quote}
Add 100 uL of blue dye reagent to every occupied well.
\end{quote}

is represented in $G_{\mathrm{sem}}$ as one semantic operation over
17 targets and normalized into 17 target-specific liquid-transfer
actions. This makes the individual executable actions explicit while
preserving their common semantic origin.

\paragraph{Dependencies and lineage.}
Normalization preserves the dependency structure established in
$G_{\mathrm{sem}}$. Every canonical action records the semantic
operation or operations from which it was derived. When a single
semantic operation is decomposed, the resulting canonical operations
retain the same source lineage.
Logical sample and material identities are likewise preserved across
the transformation. Canonicalization changes the representation and
action granularity without performing physical resource allocation.

\paragraph{Control information.}
Conditions, repetitions, endpoints, and other control information can
be retained in the \texttt{control} field of a canonical operation.
Waiting behavior is represented through the canonical
\texttt{waiting} or \texttt{incubation} actions according to the
semantics established upstream.
Canonicalization does not introduce device selection, physical well
allocation, deck placement, or backend code. These decisions belong to
the execution-grounding stage described in Appendix~D.

\paragraph{Missing information.}
Missing numerical values remain absent rather than being filled with
arbitrary defaults. Explicit runtime quantities retain their declared
runtime bindings, while derived quantities retain their formulas and
referenced runtime inputs.
This treatment preserves the distinction between information that is
specified by the protocol, information resolved during protocol
screening, and information that remains unavailable at canonicalization
time.

\subsection{Worked Representation Examples}
\label{app:ir_examples}

The following examples illustrate the correspondence between protocol
text, $G_{\mathrm{sem}}$, and $G_{\mathrm{can}}$. The JSON blocks show
selected fields only; they are not complete serialized artifacts.

\subsubsection{Direct Mapping and Quantity Normalization}

Consider the effective protocol:

\begin{quote}
Transfer 200 uL of sample\_alpha to destination\_alpha.\\
Measure absorbance of each assay reaction at 450 nm.\\
Build a standard curve and calculate each sample concentration.
\end{quote}

The first two statements correspond to executable semantic operations.
The final statement is retained as an analysis annotation.

A selected semantic excerpt is:

\begin{lstlisting}[style=json]
{
  "operation_id": "op-001",
  "operation_type": "liquid_transfer",
  "objects": [
    {
      "object_id": "sample_alpha",
      "role": "source"
    },
    {
      "object_id": "destination_alpha",
      "role": "destination"
    }
  ],
  "parameters": {
    "volume": "200 uL"
  },
  "dependencies": []
}
\end{lstlisting}

The corresponding canonical excerpt is:

\begin{lstlisting}[style=json]
{
  "normalized_operation_id": "norm-001",
  "canonical_action": "liquid_transfer",
  "derived_from_semantic_operation_ids": [
    "op-001"
  ],
  "objects": {
    "source": {
      "logical_object_id": "sample_alpha"
    },
    "destination": {
      "logical_object_id": "destination_alpha"
    }
  },
  "parameters": {
    "volume": {
      "value": 200,
      "unit": "uL",
      "source_value": "200 uL"
    }
  },
  "dependencies": []
}
\end{lstlisting}

The transformation therefore changes the representation of the
quantity and logical object roles without changing the stated transfer
semantics. The absorbance operation is represented using
\texttt{plate\_reading}, while the analysis annotation does not become
an execution node.

\subsubsection{Expansion of a Multi-Target Operation}

The second example corresponds to the multi-target operation described
in the main text:

\begin{quote}
Add 100 uL of blue dye reagent to every occupied well.
\end{quote}

For this example, the semantic representation contains 17 logical
targets:

\begin{lstlisting}[style=json]
{
  "operation_id": "op-add",
  "operation_type": "liquid_add",
  "objects": [
    {
      "object_id": "blue-dye",
      "name": "blue dye reagent",
      "role": "source"
    },
    {
      "object_id": "occupied-wells",
      "role": "destination",
      "member_count": 17
    }
  ],
  "parameters": {
    "volume_per_destination": "100 uL"
  },
  "dependencies": []
}
\end{lstlisting}

Normalization expands this collection-level semantic operation into
17 target-specific canonical actions:

\begin{equation}
v_{\mathrm{add}}
\longrightarrow
\left\{
u_{\mathrm{add},1},
\ldots,
u_{\mathrm{add},17}
\right\}.
\end{equation}

\vspace{15pt}
A selected canonical action is:

\begin{lstlisting}[style=json]
{
  "normalized_operation_id": "norm-add-01",
  "canonical_action": "liquid_transfer",
  "derived_from_semantic_operation_ids": [
    "op-add"
  ],
  "objects": {
    "source": {
      "logical_object_id": "blue-dye"
    },
    "destination": {
      "logical_object_id": "target-01"
    }
  },
  "parameters": {
    "volume": {
      "value": 100,
      "unit": "uL",
      "source_value": "100 uL"
    }
  },
  "normalization_trace": {
    "mapping_mode": "decomposed"
  }
}
\end{lstlisting}

The remaining actions correspond to the other 16 logical targets.
The target identifiers represent logical targets rather than physical
well addresses. Assignment to physical labware, devices, and locations
is deferred to the execution-grounding stage.

\section{Execution Grounding and Device Routing Details}
\label{app:execution_grounding}

\subsection{Execution-Unit Construction and Execution Modes}
\label{app:execution_units}

The Canonical Action Graph specifies the experimental actions and their
dependencies, but does not determine how these actions are executed in a
particular laboratory. Before device assignment, related canonical actions
are therefore organized into execution units. Each execution unit represents
a coherent portion of the workflow that can be considered jointly for
executor selection and resource grounding.
\vspace{-5pt}
\paragraph{Execution-unit construction.}
Execution-unit grouping follows the material-flow and dependency structure
encoded in $G_{\mathrm{can}}$. Actions may remain in the same execution unit
when they contribute to the same local experimental objective, operate on a
continuous logical material state, and can be executed consecutively without
crossing a required control or resource boundary. Table~\ref{tab:execution_unit_rules} summarizes the grouping rules. The execution-unit graph preserves the partial order of the canonical action
graph. Dependencies between actions assigned to different units induce
dependencies between the corresponding execution units. Grouping therefore
changes execution granularity without changing the experimental ordering
established by the intermediate representation.

\begin{table}[!htbp]
\centering
\small

\setlength{\tabcolsep}{4pt}
\renewcommand{\arraystretch}{1.15}

\caption{Execution-unit grouping rules.}
\label{tab:execution_unit_rules}

\begin{tabular}{
@{}
>{\raggedright\arraybackslash}p{0.25\linewidth}
>{\raggedright\arraybackslash}p{0.68\linewidth}
@{}
}

\toprule

\textbf{Criterion}
&
\centering\textbf{Grouping rule}
\arraybackslash
\\

\midrule

Local experimental objective
&
Actions that jointly accomplish the same local experimental operation
may be grouped when no intervening boundary requires independent execution.
\\[3pt]

\addlinespace[4pt]

Material continuity
&
Actions operating on the same logical sample or material state may remain
in the same unit when the intermediate state does not need to be exposed
to another unit.
\\[3pt]

\addlinespace[4pt]

Dependency structure
&
The original prerequisite relations in $G_{\mathrm{can}}$ are preserved.
Grouping does not remove a dependency required by a downstream operation.
\\[3pt]

\addlinespace[4pt]

Control boundary
&
A waiting period, incubation boundary, conditional operation, experimental
endpoint, or measurement whose result is consumed downstream may separate
execution units.
\\[3pt]

\addlinespace[4pt]

Resource continuity
&
Actions may be grouped only when a compatible realization exists for the
resulting unit. A required change of executor or physical carrier can
therefore introduce a unit boundary.
\\[3pt]

\addlinespace[4pt]

Intermediate-material use
&
When an intermediate material or state is independently consumed by another
part of the workflow, its producing and consuming operations remain separated
by an explicit dependency boundary.
\\

\bottomrule

\end{tabular}
\end{table}

\textbf{Automated execution} corresponds to selecting a compatible device
$d\in\mathcal{D}$. The execution unit subsequently enters device assignment,
physical-resource binding, backend generation, and the applicable validation
pathway.

\textbf{Human-coordinated execution} corresponds to the \texttt{HUMAN}
executor in Eq.~(3). This category includes operations performed directly
by a human as well as operations in which a human operates or supervises
an appropriate laboratory instrument. When relevant, the routing artifact
may retain device guidance for the operator; this does not constitute a
separate executor class.
Table~\ref{tab:execution_mode_outputs} summarizes the principal information
retained for the two modes.

\begin{table}[t]
\centering
\small
\setlength{\tabcolsep}{4pt}
\renewcommand{\arraystretch}{1.15}
\caption{Execution modes and corresponding routing information.}
\label{tab:execution_mode_outputs}
\begin{tabular}{
@{}
>{\raggedright\arraybackslash}p{0.20\linewidth}
>{\raggedright\arraybackslash}p{0.21\linewidth}
>{\raggedright\arraybackslash}p{0.52\linewidth}
@{}
}
\toprule
Mode & Eq.~(3) executor & Retained routing information \\

\midrule

Automated
&
$d\in\mathcal{D}$
&
Canonical-operation lineage and dependencies, compatible-device set, physical-resource bindings, generated backend artifact, and validation result.
\\

Human-coordinated
&
\texttt{HUMAN}
&
Source operation and canonical-operation context, dependencies,
human-execution instruction, relevant runtime inputs, and optional
device-class or device-model guidance.
\\

\bottomrule
\end{tabular}
\end{table}

\subsection{Device Capability Matrix}
\label{app:device_capability}

\paragraph{Meaning of device integration.}
The 45 devices reported in the main text are physically integrated laboratory
devices supported by the LabCompiler execution layer. For each integrated device,
the system provides an explicit capability description together with a device-facing
grounding interface or backend through which supported canonical actions can be
mapped to device-specific commands and checked against applicable device and
resource constraints.
The available validation pathway differs across device backends. 

\paragraph{Device capability matrix.}
Table~\ref{tab:device_matrix} lists the 45 physical devices used by the
grounding catalog. Canonical action names follow Appendix~\ref{app:canonical_actions}.
The validation pathways are abbreviated as \emph{official simulation},
\emph{offline simulation}, and \emph{contract validation}.

\begingroup
\scriptsize
\setlength{\tabcolsep}{2.3pt}
\renewcommand{\arraystretch}{1.12}
\setlength{\LTleft}{0pt}
\setlength{\LTright}{0pt}
\setlength{\LTpre}{4pt}
\setlength{\LTpost}{4pt}
\setlength{\LTcapwidth}{\textwidth}

\begin{longtable}{
@{}
>{\raggedright\arraybackslash}p{0.16\textwidth}
>{\raggedright\arraybackslash}p{0.11\textwidth}
>{\raggedright\arraybackslash}p{0.235\textwidth}
>{\raggedright\arraybackslash}p{0.13\textwidth}
>{\raggedright\arraybackslash}p{0.225\textwidth}
>{\raggedright\arraybackslash}p{0.08\textwidth}
@{}
}

\caption{
Device capability matrix. 
}
\label{tab:device_matrix}
\\

\toprule
Device &
Backend family &
Supported canonical actions &
Integration &
Key constraints &
Validation
\\
\midrule
\endfirsthead

\multicolumn{6}{@{}l}{
\textit{Table~\ref{tab:device_matrix} continued from previous page}
}
\\[2pt]

\toprule
Device &
Backend family &
Supported canonical actions &
Integration &
Key constraints &
Validation
\\
\midrule
\endhead

\midrule
\multicolumn{6}{r@{}}{\textit{Continued on next page}}
\\
\endfoot

\bottomrule
\endlastfoot

Opentrons OT-2
&
Opentrons
&
\texttt{liquid\_transfer},
\texttt{liquid\_distribution},
\texttt{liquid\_consolidation},
\texttt{liquid\_removal},
\texttt{pipette\_mixing},
\texttt{orbital\_shaking},
\texttt{temperature\_control},
\texttt{incubation},
\texttt{thermal\_cycling},
\texttt{magnetic\_separation}
&
Adapter + compiler
&
Positive liquid volumes; mixing volume up to 1000 uL;
shaking 200--3000 rpm; temperature, thermal-cycling, and magnetic-module
constraints are checked when applicable.
&
Official sim.
\\
\addlinespace

Opentrons Flex
&
Opentrons
&
\texttt{liquid\_transfer},
\texttt{liquid\_distribution},
\texttt{liquid\_consolidation},
\texttt{liquid\_removal},
\texttt{pipette\_mixing}
&
Adapter + compiler
&
Positive liquid volumes and compatible liquid-handling resources.
&
Official sim.
\\
\addlinespace

Hamilton STAR
&
PyLabRobot
&
\texttt{liquid\_transfer},
\texttt{liquid\_distribution},
\texttt{liquid\_consolidation},
\texttt{liquid\_removal},
\texttt{pipette\_mixing}
&
Adapter + compiler
&
Eight-channel liquid handling; volumes up to 1000 uL; compatible
plate, reservoir, and tube-rack resources.
&
Offline sim.
\\
\addlinespace

Hamilton Nimbus
&
PyLabRobot
&
\texttt{liquid\_transfer},
\texttt{liquid\_distribution},
\texttt{liquid\_consolidation},
\texttt{liquid\_removal},
\texttt{pipette\_mixing}
&
Adapter + compiler
&
Eight-channel liquid handling; volumes up to 1000 uL.
&
Offline sim.
\\
\addlinespace

Hamilton Vantage
&
PyLabRobot
&
\texttt{liquid\_transfer},
\texttt{liquid\_distribution},
\texttt{liquid\_consolidation},
\texttt{liquid\_removal},
\texttt{pipette\_mixing}
&
Adapter + compiler
&
Eight-channel liquid handling; volumes up to 1000 uL.
&
Offline sim.
\\
\addlinespace

Tecan EVO
&
PyLabRobot
&
\texttt{liquid\_transfer},
\texttt{liquid\_distribution},
\texttt{liquid\_consolidation},
\texttt{liquid\_removal},
\texttt{pipette\_mixing}
&
Adapter + compiler
&
Eight-channel liquid handling; volumes up to 1000 uL.
&
Offline sim.
\\
\addlinespace

Agilent BioTek Cytation 1
&
PyLabRobot
&
\texttt{plate\_reading}
&
Adapter + compiler
&
Absorbance, fluorescence, and luminescence; 96/384-well formats;
mode-specific wavelength requirements.
&
Offline sim.
\\
\addlinespace

Agilent BioTek Cytation 5
&
PyLabRobot
&
\texttt{plate\_reading}
&
Adapter + compiler
&
Absorbance, fluorescence, and luminescence; 96/384-well formats;
mode-specific wavelength requirements.
&
Offline sim.
\\
\addlinespace

Agilent BioTek Synergy HT
&
PyLabRobot
&
\texttt{plate\_reading}
&
Adapter + compiler
&
Absorbance, fluorescence, and luminescence; 96/384-well formats.
&
Offline sim.
\\
\addlinespace

Agilent BioTek Synergy H1
&
PyLabRobot
&
\texttt{plate\_reading}
&
Adapter + compiler
&
Absorbance, fluorescence, and luminescence; 96/384-well formats.
&
Offline sim.
\\
\addlinespace

BMG CLARIOstar
&
PyLabRobot
&
\texttt{plate\_reading}
&
Adapter + compiler
&
Absorbance, fluorescence, and luminescence; 96/384-well formats.
&
Offline sim.
\\
\addlinespace

Molecular Devices SpectraMax M5e
&
PyLabRobot
&
\texttt{plate\_reading}
&
Adapter + compiler
&
Absorbance, fluorescence, and luminescence; 96/384-well formats.
&
Offline sim.
\\
\addlinespace

Tecan Infinite 200 PRO
&
PyLabRobot
&
\texttt{plate\_reading}
&
Adapter + compiler
&
Absorbance, fluorescence, and luminescence; 96/384-well formats.
&
Offline sim.
\\
\addlinespace

Tecan Spark 20M
&
PyLabRobot
&
\texttt{plate\_reading}
&
Adapter + compiler
&
Absorbance, fluorescence, and luminescence; 96/384-well formats.
&
Offline sim.
\\
\addlinespace

Byonoy Absorbance 96 Automate
&
PyLabRobot
&
\texttt{plate\_reading}
&
Adapter + compiler
&
Absorbance mode; 96-well format; wavelength required.
&
Offline sim.
\\
\addlinespace

Byonoy Luminescence 96
&
PyLabRobot
&
\texttt{plate\_reading}
&
Adapter + compiler
&
Luminescence mode; 96-well format.
&
Offline sim.
\\
\addlinespace

Byonoy Luminescence 96 Automate
&
PyLabRobot
&
\texttt{plate\_reading}
&
Adapter + compiler
&
Luminescence mode; 96-well format.
&
Offline sim.
\\
\addlinespace

Molecular Devices SpectraMax 384Plus
&
PyLabRobot
&
\texttt{plate\_reading}
&
Adapter + compiler
&
Absorbance mode; 96/384-well formats; wavelength required.
&
Offline sim.
\\
\addlinespace

Cytomat 2C 425
&
PyLabRobot
&
\texttt{incubation}
&
Adapter + compiler
&
Positive duration and specified incubation temperature.
&
Offline sim.
\\
\addlinespace

Cytomat 5C
&
PyLabRobot
&
\texttt{incubation}
&
Adapter + compiler
&
Positive duration and specified incubation temperature.
&
Offline sim.
\\
\addlinespace

Cytomat 6000
&
PyLabRobot
&
\texttt{incubation}
&
Adapter + compiler
&
Positive duration and specified incubation temperature.
&
Offline sim.
\\
\addlinespace

Cytomat 6002
&
PyLabRobot
&
\texttt{incubation}
&
Adapter + compiler
&
Positive duration and specified incubation temperature.
&
Offline sim.
\\
\addlinespace

Cytomat 2C 450 Shake
&
PyLabRobot
&
\texttt{incubation},
\texttt{orbital\_shaking}
&
Adapter + compiler
&
Incubation temperature 4--50 $^\circ$C; applicable shaking parameters.
&
Offline sim.
\\
\addlinespace

Inheco Incubator Shaker
&
PyLabRobot
&
\texttt{incubation},
\texttt{orbital\_shaking}
&
Adapter + compiler
&
Duration, temperature, and shaking parameters when specified.
&
Offline sim.
\\
\addlinespace

Inheco SCILA
&
PyLabRobot
&
\texttt{incubation}
&
Adapter + compiler
&
Duration and incubation temperature.
&
Offline sim.
\\
\addlinespace

LiCONiC STX
&
PyLabRobot
&
\texttt{incubation}
&
Adapter + compiler
&
Duration and incubation temperature.
&
Offline sim.
\\
\addlinespace

Hamilton Heater Shaker
&
PyLabRobot
&
\texttt{orbital\_shaking},
\texttt{temperature\_control},
\texttt{incubation}
&
Adapter + compiler
&
Incubation temperature 25--105 $^\circ$C and applicable shaking parameters.
&
Offline sim.
\\
\addlinespace

Inheco Thermoshake
&
PyLabRobot
&
\texttt{orbital\_shaking},
\texttt{temperature\_control},
\texttt{incubation}
&
Adapter + compiler
&
Temperature, duration, and shaking parameters when specified.
&
Offline sim.
\\
\addlinespace

Inheco Thermoshake AC
&
PyLabRobot
&
\texttt{orbital\_shaking},
\texttt{temperature\_control},
\texttt{incubation}
&
Adapter + compiler
&
Temperature, duration, and shaking parameters when specified.
&
Offline sim.
\\
\addlinespace

Inheco Thermoshake RM
&
PyLabRobot
&
\texttt{orbital\_shaking},
\texttt{temperature\_control},
\texttt{incubation}
&
Adapter + compiler
&
Temperature, duration, and shaking parameters when specified.
&
Offline sim.
\\
\addlinespace

QInstruments BioShake
&
PyLabRobot
&
\texttt{orbital\_shaking},
\texttt{temperature\_control},
\texttt{incubation}
&
Adapter + compiler
&
Temperature, duration, and shaking parameters when specified.
&
Offline sim.
\\
\addlinespace

Hamilton Heater Cooler
&
PyLabRobot
&
\texttt{temperature\_control},
\texttt{incubation}
&
Adapter + compiler
&
Required temperature setpoint and applicable duration.
&
Offline sim.
\\
\addlinespace

Inheco CPAC
&
PyLabRobot
&
\texttt{temperature\_control},
\texttt{incubation}
&
Adapter + compiler
&
Required temperature setpoint and applicable duration.
&
Offline sim.
\\
\addlinespace

Hamilton Tilt Module
&
PyLabRobot
&
\texttt{plate\_tilting}
&
Adapter + compiler
&
Tilt angle required; duration and repetitions retained when specified.
&
Offline sim.
\\
\addlinespace

Inheco ODTC
&
PyLabRobot
&
\texttt{thermal\_cycling}
&
Adapter + compiler
&
At least one cycle and an explicit thermal-step program.
&
Offline sim.
\\
\addlinespace

Thermo Fisher ATC
&
PyLabRobot
&
\texttt{thermal\_cycling}
&
Adapter + compiler
&
At least one cycle and an explicit thermal-step program.
&
Offline sim.
\\
\addlinespace

Thermo Fisher ProFlex
&
PyLabRobot
&
\texttt{thermal\_cycling}
&
Adapter + compiler
&
At least one cycle and an explicit thermal-step program.
&
Offline sim.
\\
\addlinespace

AIAO Centrifuge 2030
&
Supplemental
&
\texttt{centrifugation}
&
Capability + contract
&
Duration 60--3600 s; speed 100--4200 rpm or RCF 100--2000 g;
applicable ramp constraint.
&
Contract
\\
\addlinespace

Agilent Centrifuge 773
&
Supplemental
&
\texttt{centrifugation}
&
Capability + contract
&
Speed 100--3000 rpm; duration 1--86400 s; acceleration and
deceleration 0--100\%.
&
Contract
\\
\addlinespace

AIAO Dispenser 1427
&
Supplemental
&
\texttt{reagent\_dispensing}
&
Capability + contract
&
Dispense-volume, valve, dispensing-speed, priming, and related
command constraints.
&
Contract
\\
\addlinespace

Hydro Flex 588
&
Supplemental
&
\texttt{plate\_washing}
&
Capability + contract
&
Wash volume 50--3000 uL in 50-uL increments; 1--9 cycles;
channel and wash-program constraints.
&
Contract
\\
\addlinespace

WASP Sealer 2144
&
Supplemental
&
\texttt{plate\_sealing}
&
Capability + contract
&
Sealing time 2.5--9.9 s and sealing-temperature constraints.
&
Contract
\\
\addlinespace

SP 100 700
&
Supplemental
&
\texttt{plate\_sealing},
\texttt{plate\_unsealing}
&
Capability + contract
&
Film type and applicable sealing or peeling configuration.
&
Contract
\\
\addlinespace

QPix 420 2059
&
Supplemental
&
\texttt{colony\_picking}
&
Capability + contract
&
Positive colony-pick count and applicable lid configuration.
&
Contract
\\
\addlinespace

Fluent Glovebox 2035
&
Supplemental
&
\texttt{controlled\_environment}
&
Capability + contract
&
Named environment method plus at least one declared environmental
threshold and comparison.
&
Contract
\\

\end{longtable}
\endgroup

\paragraph{Compatibility function.}
The compatibility function $\Gamma(u_i,d)$ evaluates more than
action-name membership. For an execution unit $u_i$ and device $d$,
compatibility requires all applicable checks in
Table~\ref{tab:gamma_checks} to succeed. An execution unit is compatible only if a single device can realize it as a whole; support distributed across devices is insufficient.

\begin{table}[t]
\centering
\small

\setlength{\tabcolsep}{5pt}
\renewcommand{\arraystretch}{1.15}

\caption{Checks contributing to execution-unit/device compatibility.}
\label{tab:gamma_checks}

\begin{tabular}{
@{}
>{\raggedright\arraybackslash}p{0.24\linewidth}
>{\raggedright\arraybackslash}p{0.70\linewidth}
@{}
}

\toprule

\textbf{\centering Compatibility component}
&
\textbf{\centering Requirement}
\\

\midrule

\textbf{Action compatibility}
&
Every canonical action in the execution unit must have a supported
realization on the candidate device.
\\[3pt]

\addlinespace[4pt]

\textbf{Object and resource compatibility}
&
Required logical objects, carriers, labware types, and input/output
shapes must be compatible with the device interface.
\\[3pt]

\addlinespace[4pt]

\textbf{Parameter compatibility}
&
Required parameters must be present where necessary and satisfy
the applicable unit, range, and device-capability constraints.
\\[3pt]

\addlinespace[4pt]

\textbf{Backend compatibility}
&
An applicable adapter and backend realization must exist for the
action--device combination.
\\[3pt]

\addlinespace[4pt]

\textbf{Validation compatibility}
&
The realization must admit the validation pathway associated with the
backend, including simulator validation or contract validation as applicable.
\\

\bottomrule

\end{tabular}
\end{table}

\subsection{Cross-Device Resource Binding and Human Handoffs}
\label{app:resource_binding}

\paragraph{Logical-to-physical resource binding.}
Logical material identities are preserved across execution units using the
object identifiers defined in the intermediate representation. Device
assignment does not change the identity of a sample or material. Instead,
each automated execution unit introduces a physical binding between the
logical object and the resources required by the selected device.
The binding records the logical object, its carrier resource, its physical position when applicable, and the execution-unit scope in which that binding is used. This separation permits the same logical material to remain identifiable when successive execution units are assigned to different devices.

\paragraph{Cross-device handoffs.}
When a logical object produced or modified by one automated execution unit is
consumed by a dependent unit assigned to a different device, the layout layer
records a cross-device handoff. A handoff retains the same
\texttt{logical\_object\_id} while recording the physical binding on both
sides of the transition:

\begin{minipage}{\linewidth}
\begin{lstlisting}[
    basicstyle=\ttfamily\small,
    frame=single,
    breaklines=true
]
logical_object_id
from:
    device
    carrier_resource
    position
to:
    device
    carrier_resource
    position
\end{lstlisting}
\end{minipage}

\medskip

The handoff therefore represents the resource transition required by the workflow without changing the underlying logical sample identity. It is a planning and provenance artifact used to preserve cross-device correspondence.

\paragraph{Human-coordinated handoffs.}
For execution units assigned to \texttt{HUMAN}, the workflow retains an
instruction derived from the approved protocol and the corresponding
canonical-operation context. The instruction can include the relevant
materials, parameters, expected manipulation, declared runtime inputs, and,
when appropriate, device-class or device-model guidance.
Human-coordinated units are represented explicitly in the routing artifact,
while the remaining automated units continue through device assignment and
backend generation. Dependencies from automated units to human-coordinated
operations are preserved as human prerequisites in the workflow plan. The human-routing artifact specifies the required operation and its
dependencies; it is not interpreted as an additional automated device
backend or as evidence of wet-lab completion.

\subsection{Validation Levels, Compilation Failures, and Fallback}
\label{app:execution_validation}

Automated realizations are validated according to the pathway available for
the selected backend. Table~\ref{tab:validation_levels} distinguishes the
validation levels used in this work.

\begin{table}[t]
\centering
\small

\setlength{\tabcolsep}{5pt}
\renewcommand{\arraystretch}{1.15}

\caption{Validation levels for automated realizations.}
\label{tab:validation_levels}

\begin{tabular}{
@{}
>{\raggedright\arraybackslash}p{0.23\linewidth}
>{\raggedright\arraybackslash}p{0.56\linewidth}
>{\raggedright\arraybackslash}p{0.16\linewidth}
@{}
}

\toprule

\textbf{\centering Validation level}
&
\textbf{\centering What is checked}
&
\textbf{\centering Interpretation}
\\

\midrule

\textbf{Static and device-constraint validation}
&
Action mapping, required parameters, object/resource compatibility,
device-capability constraints, backend contract, and physical-layout
consistency when applicable.
&
Required for an automated realization.
\\[3pt]

\addlinespace[4pt]

\textbf{Simulator validation}
&
Generated backend artifact is evaluated using the simulator associated
with the selected device backend.
&
Applied when a simulator is available.
\\[3pt]

\addlinespace[4pt]

\textbf{Contract validation}
&
The normalized device command is checked against the registered
device-command and parameter contract.
&
Used for backends without an applicable simulator.
\\[3pt]

\bottomrule

\end{tabular}
\end{table}

\paragraph{Static and device-constraint validation.}
Before an automated realization is accepted, the selected execution unit is
checked against the device and resource requirements used during grounding.
These checks cover the canonical-action mapping, required operation parameters,
logical-object roles, relevant device capability ranges, and resource/layout
requirements.

\paragraph{Simulator validation.}
When a simulator is available for the selected backend, the generated
device-specific realization is additionally evaluated in simulation. A
simulated realization is accepted only when the generated artifact and its
simulator result satisfy the applicable backend checks.
The simulator or contract checks are provided by the corresponding device backend and are not
constructed from LabCompiler's predicted outputs. They evaluate whether the generated realization
conforms to the execution interface and constraints of the target platform.

\paragraph{Backends without a simulator.}
For device interfaces for which no simulator is used, the corresponding
device-command contract provides the applicable validation pathway. In this
case, successful static and contract validation represents the highest
validation level used by the execution-grounding stage. Such a realization is
reported as contract-validated rather than simulator-validated.

\paragraph{Compilation failures and routing behavior.}
An automated realization is accepted only when all validation stages applicable
to its selected backend succeed. If an applicable device, resource, compilation,
or simulator check fails, that automated realization is considered
unsuccessful.
For recoverable capability or configuration cases, the affected operation can
be assigned to the human-coordinated route while unaffected automated units
remain available for compilation. The resulting workflow therefore preserves
the executable automated portions of the experiment together with explicit
human-coordinated steps.
This behavior should not be interpreted as a general automatic fallback chain
that successively tries another device and then progressively degrades through
multiple human-execution modes. Alternative automated routes follow the compatible-device checks described in Appendix~\ref{app:device_capability},
whereas the human-coordinated route represents the \texttt{HUMAN} executor in
Eq.~(3).

\section{Benchmark Construction, Data Splits, and Expert Annotation}
\label{app:benchmark}

\subsection{Data Provenance and Perturbation Construction}
\label{app:data_provenance}

\paragraph{Controlled corpus and evaluation subset.}
The complete controlled corpus contains 2,400 accepted perturbation instances, with 600 instances in each of the four categories. For each category, 100 instances are reserved for self-evolution memory construction and the remaining 500 are used exclusively for held-out evaluation. Thus, the controlled benchmark results reported in the main text are computed on 2,000 held-out evaluation protocols, while the additional 400 instances are used only for constructing the troubleshooting memory.
The protocol recovery benchmark contains two complementary subsets:
a controlled perturbation subset constructed from validated protocols and
a real-world subset curated from iGEM protocols. The two subsets were
constructed independently and serve different purposes. The controlled subset
supports category-specific analysis under a single targeted perturbation,
whereas the iGEM subset evaluates recovery from naturally occurring protocol
errors.

\paragraph{Controlled source protocols.}
Source protocols for the controlled subset were expert-selected from
SciRecipe~\citep{sun2026unleashing}. The selected protocols were used as
validated source procedures from which controlled perturbations were
constructed.

For each of the four perturbation categories reported in the main text, 600 accepted
instances were retained. Within a category, each accepted instance was constructed
from a different source protocol ID; thus, a source protocol contributes at most one
instance to a given perturbation category. The same source protocol may contribute
to different perturbation categories. However, all instances derived from the same
source protocol are assigned to the same data partition, ensuring source-protocol
disjointness between the self-evolution and held-out evaluation sets. Unique-source
counts are therefore reported per category rather than summed across categories.
The four perturbation families follow the definitions in the main text.
Table~\ref{tab:perturbation_construction} provides the corresponding
construction and quality-control rules used during benchmark generation.

\begin{table}[!htbp]
\centering
\small

\setlength{\tabcolsep}{5pt}
\renewcommand{\arraystretch}{1.15}

\caption{
Construction rules for the four controlled perturbation categories.
Each controlled instance contains one targeted perturbation.
}
\label{tab:perturbation_construction}

\begin{tabular}{
@{}
>{\raggedright\arraybackslash}p{0.20\linewidth}
>{\raggedright\arraybackslash}p{0.34\linewidth}
>{\raggedright\arraybackslash}p{0.39\linewidth}
@{}
}

\toprule

\textbf{\centering Perturbation}
&
\textbf{\centering Construction rule}
&
\textbf{\centering Quality-control constraint}
\\

\midrule

\textbf{Critical-step deletion}
&
Delete one complete operation required to establish a state used by a
downstream step.
&
All remaining source steps retain their original content and relative
order, except for necessary step renumbering. The deletion must create
the intended missing prerequisite rather than modifying unrelated content.
\\[3pt]

\addlinespace[4pt]

\textbf{Operation or parameter ambiguity}
&
Retain the underlying operation while making one execution-critical
operation or quantitative specification underdetermined, for example by
removing a numerical value, unit, or required range.
&
No additional step is inserted or deleted, the step order is unchanged,
and the target is not replaced by a different concrete reagent or
parameter.
\\[3pt]

\addlinespace[4pt]

\textbf{Parameter or reagent replacement}
&
Replace one explicit experimental parameter or reagent identity with an
incompatible alternative.
&
The underlying operation, number of steps, step order, and unrelated
parameters remain unchanged.
\\[3pt]

\addlinespace[4pt]

\textbf{Cross-step inconsistency}
&
Modify a later reference to a sample, reagent, component, quantity, or
state established by an earlier step.
&
The modified statement remains locally plausible while introducing a
cross-step inconsistency. Unrelated steps are preserved.
\\

\bottomrule

\end{tabular}
\end{table}

\paragraph{Candidate generation and quality control.}
Approximately 4,000 candidate perturbation instances were generated across
the four reported categories. Candidate construction was constrained so that
the intended modification affected one target defect while leaving unrelated
protocol content unchanged.
The candidate pool was screened using automated quality-control checks to verify
the intended perturbation category, error location, single-error constraint, and
preservation of unrelated protocol content. After automated screening and quality
control, 600 accepted instances were retained for each category.
The final controlled data therefore contain 2{,}400 accepted instances, of which 100 per category are used for troubleshooting
memory construction and 500 per category form the held-out controlled
evaluation subset.

\paragraph{Scope of the single-error constraint.}
For the controlled subset, the single-error constraint is part of the
construction procedure: each instance contains one targeted perturbation
belonging to one of the four categories above. Quality-control checks verify
that the intended perturbation is present and that unrelated protocol content
has not been substantively altered.
The iGEM subset is not generated through synthetic perturbation. Consequently,
the single-error property is established through expert curation rather than
by the perturbation generator.

\paragraph{Real-world iGEM subset.}
The real-world subset was curated from publicly available experimental
protocols produced by iGEM teams
\citep{jainarayanan2021igem,igem2023teamwikis}. Subject-matter experts
manually inspected candidate protocols, identified naturally occurring
experimental defects, assigned an error category and location, and produced
a corrected reference protocol.
To ensure a well-defined evaluation target, experts retained instances for which
a primary protocol defect and a corresponding reference correction could be
reliably identified. The resulting iGEM subset contains 259 expert-curated instances spanning the same four error categories as the controlled benchmark.

\subsection{Protocol-Disjoint Splits and Leakage Audit}
\label{app:data_splits}

For each controlled perturbation category, the 600 accepted instances were
partitioned into 100 instances for self-evolution and 500 held-out instances for
evaluation. Partitioning was performed at the source-protocol level across all four
perturbation categories. Consequently, no source protocol ID appears in both the
self-evolution and held-out evaluation sets, regardless of perturbation category.
When multiple perturbation instances are derived from the same source protocol,
all such instances are assigned to the same partition. No development split is used
for this benchmark construction.
Table~\ref{tab:benchmark_splits} summarizes the resulting data partitions.
The ``Expert audit'' column refers to the 20\% expert-reviewed portion of the
held-out controlled evaluation set described in
Appendix~\ref{app:benchmark_annotation}.

\begin{table}[t]
\centering
\small
\setlength{\tabcolsep}{4pt}
\renewcommand{\arraystretch}{1.18}
\caption{
Benchmark composition and data splits.
Unique source protocols are reported within each perturbation category;
the same source protocol may appear in different categories.
}
\label{tab:benchmark_splits}
\begin{tabular}{
@{}
>{\raggedright\arraybackslash}p{0.27\linewidth}
>{\centering\arraybackslash}p{0.10\linewidth}
>{\centering\arraybackslash}p{0.13\linewidth}
>{\centering\arraybackslash}p{0.12\linewidth}
>{\centering\arraybackslash}p{0.08\linewidth}
>{\centering\arraybackslash}p{0.10\linewidth}
>{\centering\arraybackslash}p{0.10\linewidth}
@{}
}
\toprule
Dataset / category &
Accepted &
Unique source protocols &
Self-evolution &
Dev. &
Test &
Expert audit
\\
\midrule

Critical-step deletion
&
600
&
600
&
100
&
0
&
500
&
100
\\

Operation or parameter ambiguity
&
600
&
600
&
100
&
0
&
500
&
100
\\

Parameter or reagent replacement
&
600
&
600
&
100
&
0
&
500
&
100
\\

Cross-step inconsistency
&
600
&
600
&
100
&
0
&
500
&
100
\\

\midrule

Controlled total
&
2,400
&
--\textsuperscript{a}
&
400
&
0
&
2,000
&
400
\\

iGEM real-world subset
&
259
&
259
&
0
&
0
&
259
&
259
\\

\bottomrule
\end{tabular}

\vspace{2pt}
{\footnotesize
\textsuperscript{a}Unique source counts are not summed across categories because
a source protocol may contribute to more than one perturbation category. All
instances derived from the same source protocol are assigned to the same partition,
ensuring source-protocol disjointness between the self-evolution and held-out
evaluation sets.
}
\end{table}

\paragraph{Controlled evaluation split.}
The 2,000 controlled instances reported in the main experimental tables are
the 500 held-out evaluation instances from each of the four categories, for a total of 2,000 instances. The additional 100 instances per category are used only for the
perturbation-guided memory construction described in Appendix~B and are not
part of the controlled test set.

\paragraph{Frozen self-evolution memory.}
Troubleshooting memory is constructed from the 400 controlled evolution
instances and is frozen before evaluation on the held-out controlled test
protocols. Test reference corrections, true error categories, and annotated
error locations are evaluation targets and are not used to update the frozen
memory during testing.

\paragraph{iGEM isolation.}
The iGEM subset is reserved for real-world evaluation and is completely
excluded from self-evolution and development. Specifically, iGEM protocols,
reference corrections, error categories, and error locations are not used for

\begin{itemize}
    \item troubleshooting-memory construction;
    \item prompt tuning;
    \item in-context example selection; or
    \item checkpoint or model selection.
\end{itemize}

The 259 iGEM instances are therefore evaluated independently of the controlled
self-evolution process.

\subsection{Expert Review of Benchmark Annotations}
\label{app:benchmark_annotation}

Expert review in this section concerns the construction and annotation of the
benchmark itself. It is separate from the expert validation of model-response
scores described in Appendix~G.

\paragraph{Controlled perturbation audit.}
Five reviewers participated in the expert audit of
the controlled evaluation set. All reviewers held doctoral degrees and had
approximately 3--5 years of biological laboratory experience.
Reviewers examined the sampled instances with respect to four aspects:
1. whether the assigned error category matched the introduced defect;
2. whether the annotated error location identified the intended perturbation;
3. whether the reference correction restored a scientifically coherent protocol; and
4. whether experimental content unrelated to the targeted perturbation remained unaffected.
Consistent with the quality-control procedure described in the main text, experts
reviewed 20\% of the held-out evaluation instances in each controlled error
category. This corresponds to 100 instances per category and 400 controlled test
instances in total. Reviewers were compensated at USD 5 per reviewed controlled
instance.

\paragraph{Expert curation of the iGEM subset.}
The iGEM subset was curated separately from the controlled perturbation data.
Subject-matter experts manually inspected candidate protocols and retained only
instances containing an identifiable primary protocol defect for which a
scientifically defensible corrected reference could be constructed.
The subject-matter experts involved in this curation had an average Google
Scholar h-index of 27. Experts were compensated at USD~10 per retained iGEM
instance. The resulting 259 instances were individually expert curated and
paired with corrected reference protocols.

\paragraph{Reference corrections and non-unique repairs.}
The corrected reference associated with each benchmark instance specifies an
expert-validated repair, but it is not intended to define a unique textual
realization. Scientifically equivalent repairs may differ in wording or local
presentation. Accordingly, protocol-repair evaluation is semantic rather than
based on exact string matching, as described in the main text.
For the real-world subset, experts retained cases with a clearly identifiable
target defect and a scientifically defensible reference correction, enabling
consistent evaluation of protocol recovery.

\section{Component-Wise Ablations}
\label{app:ablations}

We conduct component-wise ablations separately at different stages of
LabCompiler. Each ablation modifies only the component under study while
keeping the remaining evaluation configuration fixed. This section first
examines two components of the protocol-recovery stage: the iterative
verification loop and the self-evolved troubleshooting memory. Evaluation
metrics and scoring procedures follow Appendix~G.

\subsection{SCREENING Loop}
\label{app:ablation_verification}

This ablation evaluates the contribution of iterative screening in the
protocol firewall. We compare a \emph{one-pass} variant against the complete
iterative diagnose--repair--verify procedure. The one-pass variant performs a single diagnosis-and-repair pass and returns
the resulting repair without iterative screening feedback. The iterative
variant uses the complete verification loop, in which an unsuccessful
candidate can be revised based on screening feedback before acceptance. To isolate the verification mechanism, both variants use the same
DeepSeek-V4-Pro backbone and the same screening knowledge, without
self-evolved troubleshooting memory. We randomly sample 100 test protocols
from each of the four controlled perturbation categories, resulting in
400 evaluation instances in total. Because each category contributes the same
number of protocols, the overall score is the arithmetic mean of the four
category-level scores. Table~\ref{tab:verification_loop_ablation} reports the protocol-repair
results.

\begin{table}[!htbp]
\centering
\small
\setlength{\tabcolsep}{6pt}
\renewcommand{\arraystretch}{1.15}
\caption{
Ablation of the iterative screening loop.
One hundred test protocols are randomly sampled from each controlled
perturbation category (400 protocols in total).
$\Delta$ denotes the improvement of the iterative loop over one-pass
diagnosis and repair, in score points.
}
\label{tab:verification_loop_ablation}

\begin{tabular}{
@{}
>{\raggedright\arraybackslash}p{0.43\linewidth}
ccc
@{}
}
\toprule
Perturbation category &
One-pass repair &
Iterative repair &
$\Delta$
\\
\midrule

Critical-step deletion
&
38.1
&
39.8
&
+1.7
\\

Operation or parameter ambiguity
&
27.9
&
31.2
&
+3.3
\\

Parameter or reagent replacement
&
44.3
&
53.4
&
+9.1
\\

Cross-step inconsistency
&
53.2
&
63.6
&
+10.4
\\

\midrule

\textbf{Overall}
&
\textbf{40.9}
&
\textbf{47.0}
&
\textbf{+6.1}
\\

\bottomrule
\end{tabular}
\end{table}

The iterative screening loop improves the average repair score from
40.9 to 47.0, corresponding to a gain of 6.1 score points over one-pass
diagnosis and repair. The magnitude of the improvement varies across error
types. The gains are relatively small for critical-step deletion
($+1.7$) and operation or parameter ambiguity ($+3.3$), and are larger for
parameter or reagent replacement ($+9.1$) and cross-step inconsistency
($+10.4$). This comparison isolates the effect of iterative screening
from that of the self-evolved troubleshooting memory examined below.

\subsection{Troubleshooting Memory}
\label{app:ablation_memory}

This ablation evaluates the contribution of perturbation-guided
self-evolved troubleshooting memory to protocol recovery. The backbone
model, screening knowledge, diagnose--repair--verify procedure, and
held-out evaluation protocols are kept fixed; only the amount of
troubleshooting experience incorporated into the frozen memory is varied.
The 0-example condition uses no self-evolved troubleshooting memory. The
50-example-per-category condition uses 50 protocol-recovery examples from each
perturbation category (200 examples in total), whereas the 100-example-per-category
condition uses 100 examples from each category (400 examples in total). In all
cases, the resulting memory is frozen before evaluation and is not updated using
held-out test protocols.

\begin{figure}[t]
    \centering
    \includegraphics[width=\columnwidth]{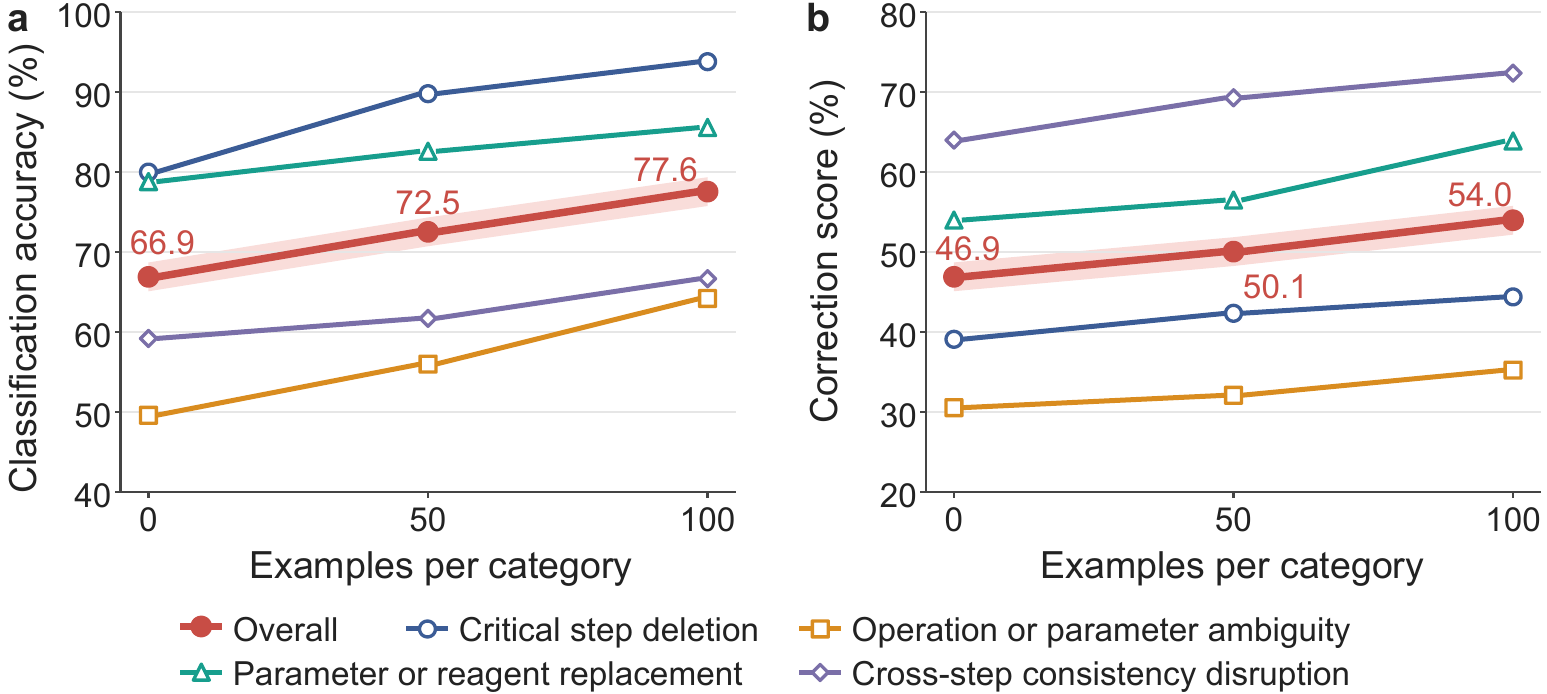}
    \caption{ Effect of self-evolved troubleshooting memory on protocol recovery.
Classification accuracy (a) and correction score (b) are reported using 0, 50, and
100 self-evolution examples per perturbation category, corresponding to 0, 200,
and 400 memory-construction examples in total, respectively. The resulting memories
are frozen before evaluation on the same held-out test set. Curves show the overall
result and the four controlled perturbation categories.}
    \label{fig:self_evolution}
\end{figure}

As shown in Figure~\ref{fig:self_evolution}, protocol-recovery performance improves progressively as additional troubleshooting experience is incorporated into the frozen memory. Overall classification accuracy increases from 66.9\% without evolved memory to 72.5\% with 50 self-evolution examples per category and 77.6\% with 100 examples per category. Relative to the no-memory condition, these correspond to absolute gains of 5.6 and 10.7 percentage points, respectively.
The correction score follows the same trend, increasing from 46.9 without evolved
memory to 50.1 and 54.0 with 50 and 100 self-evolution examples per category,
respectively, corresponding to absolute gains of 3.2 and 7.1 score points.
This ablation isolates the effect of reusable troubleshooting experience:
the underlying DeepSeek-V4-Pro backbone, screening knowledge, iterative
verification procedure, and held-out evaluation set remain unchanged
across the three memory conditions.

\subsection{Two-Level Representation}
\label{app:ablation_two_level_ir}

We evaluate whether the explicit semantic intermediate representation
contributes to the quality of the final canonical representation.
Both variants start from the same source protocols and use the same canonical action vocabulary and normalization procedure.
In the full two-level variant, the protocol is first converted into the Semantic Experiment Graph \(G_{\mathrm{sem}}\) and subsequently normalized into \(G_{\mathrm{can}}\). In the direct-to-canonical ablation, the explicit \(G_{\mathrm{sem}}\) representation is removed and canonical actions are generated directly from the protocol using the same canonical action vocabulary.
We use the same 50 source protocols and representation-evaluation protocol
described in Appendix~G. Results are shown in
Table~\ref{tab:ablation_two_level_ir}.

\begin{table*}[t]
\centering
\small
\setlength{\tabcolsep}{5pt}
\renewcommand{\arraystretch}{1.15}
\caption{
Ablation of the two-level intermediate representation.
Higher values are better. $\Delta$ denotes the improvement of the two-level representation over direct-to-canonical mapping.
}
\label{tab:ablation_two_level_ir}

\begin{tabular}{lcccccc}
\toprule
&
\multicolumn{3}{c}{Fidelity} &
\multicolumn{3}{c}{Stability}
\\
\cmidrule(lr){2-4}
\cmidrule(lr){5-7}

Representation &
Operation &
Detail &
Relation &
Operation &
Detail &
Relation
\\
\midrule

Direct-to-canonical
(w/o $G_{\mathrm{sem}}$)
&
88.6 &
87.6 &
85.4 &
89.7 &
86.5 &
83.2
\\

Two-level
($G_{\mathrm{sem}}\rightarrow G_{\mathrm{can}}$)
&
\textbf{96.4} &
\textbf{96.2} &
\textbf{94.6} &
\textbf{95.8} &
\textbf{97.3} &
\textbf{98.2}
\\

\midrule

$\Delta$
&
\textbf{+7.8} &
\textbf{+8.6} &
\textbf{+9.2} &
\textbf{+6.1} &
\textbf{+10.8} &
\textbf{+15.0}
\\

\bottomrule
\end{tabular}
\end{table*}

Removing $G_{\mathrm{sem}}$ degrades all six representation metrics.
The largest reduction occurs in relation stability, where the
two-level representation improves the score by 15.0 points.
The results indicate that explicitly reconstructing protocol-level
semantics before canonicalization is particularly important for preserving
details and relations across semantically equivalent rewrites.

\subsection{Execution-Unit Grouping}
\label{app:ablation_execution_unit}

Table~\ref{tab:ablation_execution_unit} shows the contribution of execution-unit grouping. The full variant groups related canonical actions according to their
material flow and dependencies before capability matching. In the ablated
variant, each canonical action is treated as an independent singleton
execution unit and routed separately.
Both variants use the same canonical representations, device capability
catalog, compatibility checks, and validation procedure. We evaluate them
on the same 50 protocols sampled from the execution-grounding evaluation
set. Metric definitions follow Appendix~G.
Execution-unit grouping improves route accuracy from 61.2\% to 72.9\% and device
executability from 49.5\% to 59.6\%, corresponding to gains of 11.7 and 10.1
percentage points, respectively. This comparison shows that capability grounding
benefits from considering related actions jointly rather than routing each canonical
action in isolation.

\begin{table}[t]
\centering
\small
\setlength{\tabcolsep}{8pt}
\renewcommand{\arraystretch}{1.15}
\caption{
Ablation of execution-unit grouping on the same 50 execution protocols.
$\Delta$ denotes the improvement obtained by grouping related actions
before capability matching.
}
\label{tab:ablation_execution_unit}

\begin{tabular}{lccc}
\toprule
Routing granularity &
Execution-unit grouping &
Route accuracy (\%) &
Device executability (\%)
\\
\midrule

Action-wise routing
&
No
&
61.2
&
49.5
\\

Execution-unit routing
&
Yes
&
\textbf{72.9}
&
\textbf{59.6}
\\

\midrule

$\Delta$
&
--
&
\textbf{+11.7}
&
\textbf{+10.1}
\\

\bottomrule
\end{tabular}
\end{table}

\section{Evaluation Protocols, Expert--Judge Agreement, and Statistical Details}
\label{app:evaluation}

\subsection{Experimental Configurations and Stage-Wise Evaluation Inputs}
\label{app:experimental_configuration}

We evaluate LabCompiler at three distinct interfaces:
(i) protocol recovery, which evaluates the protocol firewall;
(ii) device-agnostic representation, which evaluates the transformation from a source protocol to the canonical intermediate representation; and
(iii) execution grounding, which evaluates routing, device executability,
and device-specific realization.
The evaluation setup is intentionally stage dependent. Protocol recovery is
evaluated independently from downstream compilation. Representation fidelity is
evaluated directly against the source protocol, whereas the execution experiment
begins from the protocol input and allows upstream representation errors to
propagate into downstream grounding.

\paragraph{Methods and baselines.}
The protocol-recovery experiments compare direct LLM baselines, a coding-agent
baseline, LabCompiler, and the self-evolved LabCompiler variant. The direct
LLM baselines are MiniMax-M3, Qwen3.8-Max, DeepSeek-V4-Pro,
Gemini-3.1-Pro, Kimi-K3, GPT-5.6 Sol, and Claude Opus 5.
Codex with GPT-5.6 Sol is used as the coding-agent baseline.

Both LabCompiler and LabCompiler* use DeepSeek-V4-Pro as the underlying
language model. LabCompiler uses the protocol-screening knowledge available
to the firewall but does not use evolved troubleshooting memory.
LabCompiler* uses the same backbone and screening configuration together with
a frozen self-evolved troubleshooting memory constructed from 100
self-evolution examples per perturbation category.

\subsection{Scoring Rubrics and Metric Denominators}
\label{app:scoring}

The reported metrics operate at different evaluation units (Table~\ref{tab:metric_summary} ). 
For metrics with multiple evaluation instances per source protocol (e.g., execution units within protocols or rewrites derived from a protocol), scores are first aggregated within each protocol using the corresponding reference instances as denominators, and then averaged across protocols.

\begin{table*}[!htbp]
\centering
\small

\setlength{\tabcolsep}{3.5pt}
\renewcommand{\arraystretch}{1.12}

\caption{Summary of evaluation metrics, within-protocol denominators, and final aggregation procedures.}
\label{tab:metric_summary}

\begin{tabular}{
@{}
>{\raggedright\arraybackslash}p{0.17\textwidth}
>{\raggedright\arraybackslash}p{0.13\textwidth}
>{\raggedright\arraybackslash}p{0.14\textwidth}
>{\raggedright\arraybackslash}p{0.24\textwidth}
>{\raggedright\arraybackslash}p{0.24\textwidth}
@{}
}

\toprule

\multicolumn{1}{c}{\textbf{Metric}}
&
\multicolumn{1}{c}{\makecell{\textbf{Evaluation}\\\textbf{unit}}}
&
\multicolumn{1}{c}{\makecell{\textbf{Raw}\\\textbf{scale}}}
&
\multicolumn{1}{c}{\makecell{\textbf{Within-protocol}\\\textbf{denominator}}}
&
\multicolumn{1}{c}{\makecell{\textbf{Final}\\\textbf{aggregation}}}
\\

\midrule

\textbf{Classification accuracy}
&
Protocol
&
0/1
&
All evaluated controlled protocols
&
Mean correctness over protocols
\\

\addlinespace[3pt]

\textbf{Repair score}
&
Protocol
&
0--2 per judge
&
All protocols with paired judge scores
&
Two-judge mean, linearly mapped to 0--100
\\

\addlinespace[3pt]

\textbf{IR fidelity}
&
Protocol
&
0--5 per dimension
&
50 original protocols
&
Mean across protocols
\\

\addlinespace[3pt]

\textbf{IR stability}
&
Rewrite
&
0--5 per dimension
&
100 rewrites per transformation
&
Average within source protocol, then across source protocols
\\

\addlinespace[3pt]

\textbf{Route accuracy}
&
Execution unit
&
0/1
&
Reference execution units in each protocol
&
Protocol-level accuracy, then mean across protocols
\\

\addlinespace[3pt]

\textbf{Device executability}
&
Automated execution unit
&
0/1
&
Reference automated execution units in each protocol
&
Protocol-level success rate, then mean across protocols
\\

\addlinespace[3pt]

\textbf{Code fidelity}
&
Automated execution unit
&
Three 0--5 dimensions
&
Reference automated execution units in each protocol
&
Node score $\rightarrow$ protocol mean $\rightarrow$ mean across protocols
\\

\bottomrule

\end{tabular}
\end{table*}

\subsubsection{Protocol-Recovery Metrics}

\paragraph{Classification accuracy.}
Each controlled instance belongs to one of the four perturbation categories
defined in Section~4 and Appendix~E. A prediction receives a score of one only
when the predicted category exactly matches the reference category.
For $N$ evaluated protocols:

\begin{equation}
\mathrm{Acc}
=
\frac{1}{N}
\sum_{i=1}^{N}
\mathbb{I}\!\left[
\hat{y}_i = y_i
\right].
\end{equation}

The controlled evaluation contains 500 test protocols in each of the four
categories. Consequently, the pooled protocol-level mean over all 2,000
instances is numerically identical to the macro-average of the four
category-level accuracies.

\Needspace{12\baselineskip}

\paragraph{Repair score.}
Protocol repair is scored on the 0--2 scale defined in the main text:

\begin{itemize}
    \item the target defect is fully resolved and the repair is
    scientifically consistent with the reference;
    \item the repair is partially correct but contains a minor
    omission or inaccuracy;
    \item the defect remains unresolved or the repair introduces
    a substantive error.
\end{itemize}

Both judge calls succeeded for the reported experiments, so no protocol was removed because of judge failure.

\subsubsection{Representation Fidelity and Stability}

Both representation fidelity and representation stability are evaluated along
three dimensions:

\begin{itemize}
    \item \textbf{Operation coverage}: whether required experimental operations
    are represented without unsupported additions;
    \item \textbf{Detail preservation}: whether execution-relevant materials,
    quantities, parameters, and conditions are retained;
    \item \textbf{Relation preservation}: whether procedural, material-flow,
    and dependency relations among operations are preserved.
\end{itemize}

Each dimension is scored on a 0--5 ordinal scale by GPT-5.5 and
Gemini-3.1-Pro. A score of 5 indicates no identified representation error;
a score of 4 corresponds to only minor issues, while a critical semantic
failure constrains the score to 3 or below.
For dimension $k$, the paired judge score for protocol $i$ is

\begin{equation}
\bar{s}_{i,k}
=
\frac{
s_{i,k}^{(1)} + s_{i,k}^{(2)}
}{2}.
\end{equation}

The normalized percentage score is obtained by dividing by 5 and multiplying
by 100.

\paragraph{Representation fidelity.}
Fidelity is evaluated on 50 original source protocols. For each protocol,
the generated $G_{\mathrm{can}}$ is evaluated against the corresponding
source protocol along operation coverage, detail preservation, and relation
preservation. All representation-fidelity artifacts are also individually
reviewed by biological-domain experts to verify semantic correspondence with
the source protocol and preservation of execution-relevant information.

\paragraph{Representation stability.}
The semantics-preserving rewrite set was designed by five biological-domain
experts based on common variations in how laboratory procedures are expressed
in practice. Five transformation types are considered: content reordering,
word replacement, context expansion, structure rewrite, and unit conversion.

For each of the 50 source protocols, ten rewrites were constructed, with two
instances for each transformation type. The rewrites were expert-checked to
ensure that the experimental operations, parameters, conditions, and
procedural relationships remained semantically equivalent to those in the
corresponding source protocol. This yields 100 rewrite instances per
transformation and 500 rewrite instances in total.
The reference for stability is the canonical representation generated from
the corresponding original protocol. A rewritten protocol is processed through
the same pipeline and its resulting $G_{\mathrm{can}}$ is compared with this
reference representation.
For each transformation, the two rewrites associated with a source protocol
are first averaged within that protocol. The resulting protocol-level scores
are then averaged across the 50 source protocols. This gives equal weight to
each original protocol rather than to individual rewrite instances.

\subsubsection{Execution-Grounding Metrics}
These protocols were manually reviewed by biological-domain experts to establish the execution
ground truth before evaluation. For each protocol, experts decomposed the workflow into reference
execution units and annotated the expected executor type together with the applicable
device-capability requirements for automated execution. These reference requirements were
determined from the experimental requirements of each execution unit and the available laboratory
device capabilities.
The resulting ground-truth set contains 643 reference execution units, including 376 units annotated
for automated device execution and 267 units annotated for HUMAN execution. These expert
annotations serve as the reference targets for routing and execution-grounding evaluation.

\paragraph{Route accuracy.}
Route accuracy is computed at the execution-unit level using expert-reviewed reference requirements. A prediction is counted as correct when it satisfies the executor requirement for that unit; human-coordinated units require the HUMAN route. For automated units, compatibility is determined by the device-capability checks in Appendix~D.2.

For protocol $p$ with $N_p$ reference execution units,

\begin{equation}
\mathrm{RouteAcc}_p
=
\frac{
\#\{\text{reference units with correct executor assignment}\}
}{
N_p
}.
\end{equation}

The reported route accuracy is the arithmetic mean of
$\mathrm{RouteAcc}_p$ across the 100 protocols. The denominator is therefore
defined by reference execution units rather than by the number of units
predicted by a method.

\paragraph{Device executability.}
Device executability is evaluated only on the 376 reference execution units
annotated for Automated execution. A unit is counted as executable only when the method routes it to Automated execution and the
resulting realization passes the validation pathway provided by the corresponding device backend.
A reference Automated unit receives a score of zero when any of the following
occurs:

\begin{itemize}
    \item the unit is routed to HUMAN;
    \item no device-specific realization is produced;
    \item compilation fails;
    \item applicable device or resource constraints fail; or
    \item simulator validation fails when a simulator is available.
\end{itemize}

For backends without an applicable simulator, successful static and
device-contract validation constitutes a successful executability result, as
described in Appendix~D.
For protocol $p$, device executability is

\begin{equation}
\mathrm{Exec}_p
=
\frac{
\#\{\text{successful reference Automated units}\}
}{
\#\{\text{reference Automated units}\}
}.
\end{equation}

The reported result is the mean of the protocol-level executability scores.

\paragraph{Code fidelity.}
Code fidelity evaluates whether the device-specific realization preserves the
intended experimental procedure. It is evaluated over all reference Automated
units, rather than only units that pass executability validation.
If no executable realization is generated for a reference Automated unit, the unit receives a fidelity score of zero and remains in the denominator.
For a generated realization, each LLM judge evaluates three dimensions on a
0--5 scale:

\begin{itemize}
    \item \textbf{Action correctness};
    \item \textbf{Experimental constraint preservation};
    \item \textbf{Logic preservation}.
\end{itemize}

The code-fidelity judge receives the experimental goal, protocol text,
generated device-specific realization, the reference execution-unit action and
summary, and the scoring rubric. The judge is not asked to re-evaluate
routing or simulator success.
Let $a_i^{(j)}$, $c_i^{(j)}$, and $l_i^{(j)}$ denote the three dimension
scores from judge $j$. The two judges are first averaged within each dimension:

\begin{equation}
\bar{a}_i
=
\frac{a_i^{(1)}+a_i^{(2)}}{2},
\qquad
\bar{c}_i
=
\frac{c_i^{(1)}+c_i^{(2)}}{2},
\qquad
\bar{l}_i
=
\frac{l_i^{(1)}+l_i^{(2)}}{2}.
\end{equation}

The normalized node-level fidelity score is then

\begin{equation}
\mathrm{Fidelity}_i
=
100
\frac{
\bar{a}_i+\bar{c}_i+\bar{l}_i
}{15}.
\end{equation}

Node-level scores are averaged within each protocol, and the reported code
fidelity is the arithmetic mean of the resulting protocol-level scores.
All LLM-judge calls used for the reported code-fidelity results completed
successfully.

\subsection{Expert Validation of LLM-Judge Scores}
\label{app:expert_judge}

Expert validation in this section concerns agreement with the LLM-judge scoring
and is distinct from the expert review of benchmark annotations described in
Appendix~E and from the expert review of representation artifacts. All
representation-fidelity artifacts are individually reviewed by domain experts
against their corresponding source protocols to verify semantic correspondence
and preservation of execution-relevant information.

Separately, to assess the reliability of the LLM-based evaluation scores, 30\% of
model responses were sampled within each model--category stratum. This score-level
validation covers protocol repair, IR fidelity, IR stability, and code fidelity
across the corresponding evaluation datasets (Table~\ref{tab:expert_judge_validation} ). The 30\% sampling therefore refers
to the independent audit of LLM-judge scores rather than to the expert verification
of the representation-fidelity artifacts themselves.

Eight biological-domain experts participated in this score-level validation. Each
expert had 3--5 years of relevant biological research or laboratory experience.
Every sampled response was independently scored by two experts. Experts were
blinded to both the identity of the model that produced the response and the scores
assigned by the LLM judges. Experts used the same task-specific rubrics as the LLM
judges: the 0--2 protocol-repair rubric, the 0--5 representation rubric, and the
three 0--5 dimensions used for code fidelity.
For each task, expert scores were aggregated using the same scale and
normalization convention as the corresponding LLM-judge metric.
Across the audited tasks, the difference between aggregate expert scores and
aggregate LLM-judge scores ranged from $-1.6$ to $+1.8$ percentage points.
The expert validation is used to assess the consistency of the judge-based
semantic metrics rather than to replace the primary LLM-judge scores reported
in the main experiments.

\begin{table}[t]
\centering
\small
\setlength{\tabcolsep}{5pt}
\renewcommand{\arraystretch}{1.15}
\caption{
Coverage of independent expert auditing of LLM-judge scores.
The reported fractions refer only to score-level validation and are separate
from the expert verification of representation-fidelity artifacts described above.
Each sampled response was evaluated independently by two blinded domain experts
using the same rubric as the corresponding LLM judges.
}
\label{tab:expert_judge_validation}

\begin{tabular}{
@{}
>{\raggedright\arraybackslash}p{0.34\linewidth}
>{\centering\arraybackslash}p{0.25\linewidth}
>{\centering\arraybackslash}p{0.27\linewidth}
@{}
}
\toprule
Task &
Score-audit fraction &
Independent expert ratings per response
\\
\midrule

Protocol repair
&
30\%
&
2
\\

IR fidelity
&
30\%
&
2
\\

IR stability
&
30\%
&
2
\\

Code fidelity
&
30\%
&
2
\\

\bottomrule
\end{tabular}
\end{table}

\subsection{Reporting Conventions and Statistical Details}
\label{app:reporting}

Table~\ref{tab:evaluation_sample_sizes} summarizes the principal evaluation
populations used in the main experiments.
Retries caused by transport or structured-output failures are treated as
execution retries of the same benchmark instance rather than as independent
experimental repetitions.

\begin{table*}[t]
\centering
\small

\setlength{\tabcolsep}{5pt}
\renewcommand{\arraystretch}{1.08}

\caption{Evaluation populations for the main experiments.}
\label{tab:evaluation_sample_sizes}

\begin{tabular}{
@{}
>{\raggedright\arraybackslash}p{0.25\textwidth}
>{\centering\arraybackslash}p{0.15\textwidth}
>{\centering\arraybackslash}p{0.18\textwidth}
>{\raggedright\arraybackslash}p{0.32\textwidth}
@{}
}

\toprule

\multicolumn{1}{c}{\textbf{Evaluation}}
&
\multicolumn{1}{c}{\textbf{Source protocols}}
&
\multicolumn{1}{c}{\textbf{Evaluation instances}}
&
\multicolumn{1}{c}{\textbf{Notes}}
\\

\midrule

\textbf{Controlled protocol recovery}
&
2,000
&
2,000
&
500 protocols in each of four perturbation categories
\\

\addlinespace[4pt]

\textbf{iGEM protocol recovery}
&
259
&
259
&
Independent real-world held-out subset
\\

\addlinespace[4pt]

\textbf{Representation fidelity}
&
50
&
50
&
Source protocol $\rightarrow G_{\mathrm{can}}$
\\

\addlinespace[4pt]

\textbf{Representation stability}
&
50
&
500 rewrites
&
10 rewrites per source protocol; 100 instances for each of five
transformations
\\

\addlinespace[4pt]

\textbf{Execution grounding}
&
100
&
643 execution units
&
376 Automated and 267 HUMAN reference units
\\

\bottomrule

\end{tabular}
\end{table*}

\paragraph{Averaging conventions.}
Controlled protocol-recovery results are computed over the complete set of
2,000 test protocols. Because the four controlled categories are balanced
at 500 protocols each, pooled protocol-level averaging and category-level
macro averaging yield the same overall value.
Representation fidelity is averaged across the 50 source protocols.
For stability, rewrite scores are first averaged within each source protocol
for a given transformation and then averaged across source protocols.
Execution metrics are first computed within each protocol and are then
averaged across the 100 protocols. This convention gives equal weight to each
protocol rather than weighting protocols in proportion to their number of
execution units.

\paragraph{Rounding.}
Internal metric summaries retain two decimal places. Values displayed in the
main tables and figures are rounded to the precision shown in the corresponding
presentation, typically one decimal place for the headline comparison plots
and tables. All normalization to a 0--100 scale is performed before display
rounding.

\section{Additional Results on Lab-Bench}
\label{app:additional_results}
We additionally evaluate LabCompiler on the public ProtocolQA subset of
LAB-Bench~\citep{laurent2024lab}. The evaluation contains 108 public
ProtocolQA questions and uses the benchmark's Accuracy, Precision, and
Coverage metrics (Table~\ref{tab:labbench_results}). LabCompiler obtains 0.86 accuracy and 0.88 precision, compared with
0.83 and 0.85 for GPT-5.6 Sol. 

\begin{table}[t]
\centering
\small
\setlength{\tabcolsep}{7pt}
\renewcommand{\arraystretch}{1.12}
\caption{
Results on the 108-question public ProtocolQA subset of LAB-Bench.
These metrics follow the LAB-Bench task definition and should not be
directly compared with the open-ended protocol-recovery metrics used in
our main benchmark.
}
\label{tab:labbench_results}

\begin{tabular}{lccc}
\toprule
Method &
Accuracy &
Precision &
Coverage
\\
\midrule

Human
& 0.79 & 0.87 & 0.91 \\

Claude 3.5 Sonnet
& 0.48 & 0.66 & 0.73 \\

Claude 3 Opus
& 0.52 & 0.62 & 0.84 \\

Gemini 1.5 Pro
& 0.47 & 0.58 & 0.81 \\

GPT-4o
& 0.53 & 0.56 & 0.95 \\

GPT-4 Turbo
& 0.37 & 0.59 & 0.62 \\

Claude 3 Haiku
& 0.45 & 0.51 & 0.87 \\

Llama-3-70B-Instruct
& 0.44 & 0.49 & 0.90 \\

GLM-5.3
& 0.73 & 0.75 & 0.98 \\

DeepSeek-V4-Pro
& 0.80 & 0.84 & 0.94 \\

Kimi-K3
& 0.82 & 0.84 & 0.98 \\

GPT-5.6 Sol
& 0.83 & 0.85 & 0.98 \\

\textbf{LabCompiler}
& \textbf{0.86}
& \textbf{0.88}
& \textbf{0.98}
\\

\bottomrule
\end{tabular}
\end{table}

\end{document}